\documentclass[sigplan,nonacm,prologue,table]{acmart}
\usepackage{hyperref}
\usepackage{gensymb}
\usepackage[normalem]{ulem}
\usepackage{amsmath,amsfonts}
\usepackage{algorithmic}
\usepackage{graphicx}
\usepackage{textcomp}
\usepackage{enumitem}
\usepackage{multirow}
\usepackage{listings}
\usepackage{subcaption}
\usepackage{soul}
\usepackage{comment}
\usepackage{tikz}
\usepackage{xspace}
\usepackage{booktabs}
\usepackage{multirow}
\usepackage{array}
\usepackage{bm}
\usepackage{xcolor} 
\hypersetup{
    unicode=true,
    bookmarksnumbered=true,
    colorlinks=true,
    linkcolor={red!70!black},
    citecolor={red!70!black},
    urlcolor={blue!70!black},
    pdfborder={0 0 0}
}
\usepackage[capitalize,nameinlink]{cleveref}
\usepackage{flushend}
\usepackage{pifont}

\newcommand{\myparagraph}[1]{\vspace{\smallskipamount}\noindent\textbf{#1.\xspace}}

\newcommand{\myparagraphemph}[1]{\vspace{\smallskipamount}\noindent\emph{#1.\xspace}}
\newcommand{\eg}{\emph{e.g.}\xspace}

\newcommand{\ie}{\emph{i.e.}\xspace}

\newcommand{\vs}{\emph{vs.}\xspace}
\newcommand*{\rom}[1]{\uppercase\expandafter{\romannumeral #1\relax}}

\newcommand{\sarrow}{\bm{$\rightarrow$}}
\newcommand{\marrow}{\tikz[baseline=-0.5ex]\draw[->, line width=1pt] (0,0) -- (0.2,0.2);}
\newcommand{\farrow}{\bm{$\uparrow$}}

\newif\ifshowcomment
\showcommenttrue

\ifshowcomment
    \newcommand{\inigo}[1]{{\color{blue}[IG: #1]}}
    \newcommand{\chaojie}[1]{{\color{magenta}[CZ: #1]}}
    \newcommand{\jovan}[1]{{\color{olive}[JS: #1]}} 
    \newcommand{\todo}[1]{{\color{red}[TODO: #1]}}
\else
    \newcommand{\inigo}[1]{\ignorespaces}
    \newcommand{\chaojie}[1]{\ignorespaces}
    \newcommand{\jovan}[1]{\ignorespaces}
    \newcommand{\todo}[1]{\ignorespaces}
\fi

\title{Rearchitecting Datacenter Lifecycle for AI: \\A TCO-Driven Framework \vspace{15pt}}

\author[]{Jovan Stojkovic}
\affiliation{University of Illinois Urbana-Champaign, USA\country{}}
\email{jovans2@illinois.edu}

\author[]{Chaojie Zhang}
\affiliation{Microsoft Azure Research\\\country{Redmond, USA}}
\email{chaojiezhang@microsoft.com}

\author[]{Íñigo Goiri}
\affiliation{Microsoft Azure Research\\\country{Redmond, USA}}
\email{inigog@microsoft.com}

\author[]{Ricardo Bianchini}
\affiliation{Microsoft Azure \\\country{Redmond, USA}}
\email{ricardob@microsoft.com}

\author[]{~\vspace{10pt}}

\date{}

\renewcommand\footnotetextcopyrightpermission[1]{}

\begin{document}

\begin{abstract}

The rapid rise of large language models (LLMs) has driven an enormous demand for AI inference infrastructure, mainly powered by high-end GPUs.
While these accelerators offer immense computational power, they incur high capital and operational costs due to frequent upgrades, dense power consumption, and cooling demands, making total cost of ownership (TCO)
for AI datacenters
a critical concern for cloud providers.

Unfortunately, traditional datacenter lifecycle management (designed for general-purpose workloads) struggles to keep pace with AI's fast-evolving models, rising resource needs, and diverse hardware profiles.
In this paper, we rethink the AI datacenter lifecycle scheme across three stages:
building, hardware refresh, and operation.
We show how design choices in power, cooling, and networking \emph{provisioning} impact long-term TCO.
We also explore \emph{refresh} strategies aligned with hardware trends.
Finally, we use \emph{operation} software optimizations to reduce cost.

While these optimizations at each stage yield benefits, unlocking the full potential requires rethinking the entire lifecycle.
Thus, we present a holistic lifecycle management framework that coordinates and co-optimizes decisions across all three stages, accounting for workload dynamics, hardware evolution, and system aging.
Our system reduces the TCO by up to 40\% over traditional approaches.
Using our framework we provide guidelines on how to manage AI datacenter lifecycle for the future.
\end{abstract}

\maketitle

\section{Introduction}

Generative LLMs are reshaping industries, from  education~\cite{education} and healthcare~\cite{llmHealth} to software development~\cite{copilot} and scientific research~\cite{zhang2024comprehensivesurveyscientificlarge}.
Their rapid adoption is driven by the ability to perform complex reasoning, summarization, and interactive tasks with minimal supervision, creating unprecedented demand for scalable AI inference infrastructure~\cite{statChat}.

Modern LLM inference relies on high-end GPUs (\eg{}, NVIDIA's A100~\cite{a100} and H100~\cite{h100}) which offer strong performance but come with steep financial and infrastructure costs.
For example, a single NVIDIA DGX H100 server can exceed \$200,000~\cite{h100Cost} and draw up to 10.2kW~\cite{splitwise,stojkovic2024dynamollm}, pushing demands for power and cooling capacity far beyond those of traditional CPU servers~\cite{h100-cooling}.
To support these workloads, cloud providers have built specialized datacenters for high-throughput inference, making AI-serving one of the most resource-intensive and costly datacenter operations~\cite{aidccost-mckinsey}.

Researchers have proposed software and hardware techniques that improve the performance~\cite{orca,vllm,flashattention,spotserve,pets,splitwise,alizadeh2024llm,alisa} or energy-efficiency~\cite{stojkovic2024dynamollm,stojkovic2025tapas,wordstowatts,towardsGreen} of LLM inference clusters.
However, for providers, the key challenge is minimizing the TCO over the datacenter lifecycle, spanning CapEx (\eg{}, infrastructure build-out) and OpEx (\eg{}, energy) while meeting performance expectations from their users.

Traditional datacenter practices (\eg{}, regular refresh cycles~\cite{actTool} and conservative provisioning~\cite{smartoclock,smoothOperator,flexDatacenter,thunderbolt}) fall short for AI workloads, where models grow rapidly in complexity and scale~\cite{EpochNotableModels2024}, hardware has higher cost and infrastructure demands~\cite{emrCost,h100Cost}, and inference is highly sensitive to latency and model quality~\cite{stojkovic2024dynamollm,stojkovic2025tapas}.

\myparagraph{Our Work} 
To address this challenge, we first break down datacenter lifecycle into stages:
\emph{build}, \emph{IT provisioning}, and \emph{operation},
each with distinct costs and optimization opportunities. 
The \emph{build} stage covers initial infrastructure setup, including provisioning decisions about power topology (\eg{}, flat~\cite{bianchini2024datacenter} \vs{} hierarchical~\cite{flexDatacenter,powerBarroso}), cooling technology (\eg{}, air \vs{} liquid~\cite{google_liquid_cool, hybridcooling}), 
and network configurations (\eg, NVLink~\cite{nvlink} \vs{} Ethernet). 
The \emph{IT provisioning} stage governs when and how to decommission and upgrade hardware. 
The \emph{operation} stage focuses on runtime management:
workload placement, scheduling, and software optimizations.

We introduce a framework that rethinks and rearchitects the entire lifecycle for AI datacenters, by exploring alternative strategies across all stages and identifies the most cost-effective combination.
In \emph{build}, we compare emerging infrastructure designs to understand and balance long-term scalability, efficiency, and performance. 
In \emph{IT provisioning}, we evaluate when to adopt new hardware and retire old systems, assessing the impact on TCO given AI’s distinct model and hardware characteristics.  
In \emph{operation}, we assess the impact of software techniques (\eg{}, model migration, LLM inference disaggregation, and workload scheduling) over lifecycle TCO.

These stages are interdependent and our framework captures cross-stage interactions to support lifecycle-aware decisions.
Leveraging workload growth trends, hardware road maps, and cost models, it projects future scenarios and selects strategies that satisfy performance, fault-tolerance, and accuracy requirements.
For example, investing in a larger power-sharing domain is more expensive at \emph{build} time but provides flexibility for future \emph{IT provisioning} and improves utilization during \emph{operation}.

We build our TCO model using open-source LLMs, public hardware specs, and detailed cost data from public sources. 
Stage-specific optimizations reduce TCO by 15\% (\emph{build}), 23\% (\emph{IT provisioning}), and 19\% (\emph{operation}).
Our cross-stage strategy achieves up to a 40\% reduction.
Looking ahead, we identify emerging cross-stage opportunities and provide guidelines for adapting AI datacenter lifecycle management to future model and hardware trajectories.

\myparagraph{Summary}
This paper makes three main contributions:
\begin{itemize}[leftmargin=*]
\item Characterization of LLM workload and GPU performance-power interactions across datacenter lifecycle stages.
\item Evaluation of stage-wise strategies that improve efficiency and cross-generation compatibility.
\item The first-of-its-kind framework for lifecycle-aware, cross-stage optimization of AI datacenters.
\end{itemize}

\begin{figure}[t]
    \vspace{-6mm}
    \centering
    \includegraphics[width=\columnwidth]{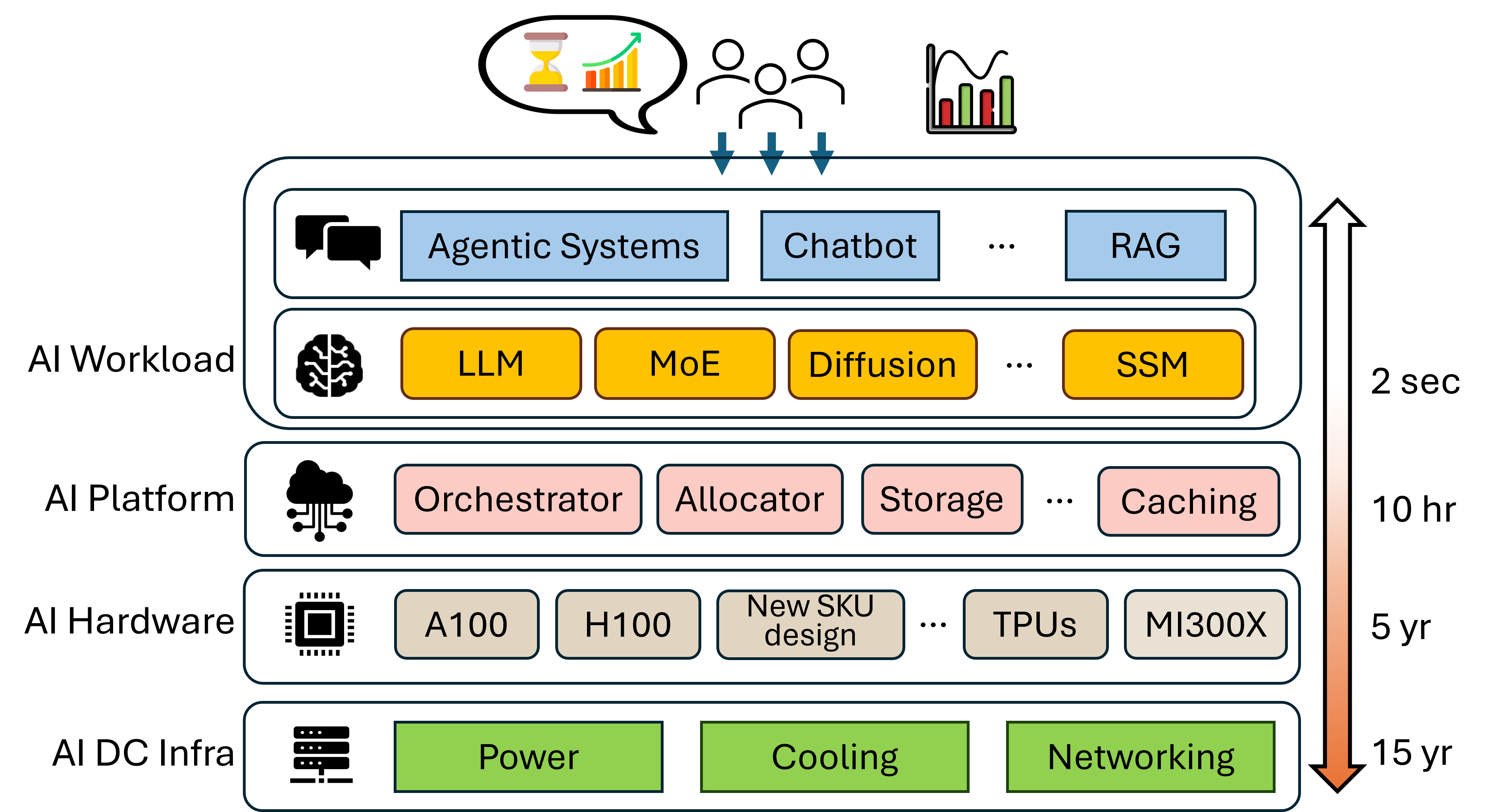}
    \vspace{-4mm}
    \caption{Hosting AI workloads from models to hardware and supporting datacenter infrastructure.
    }
    \label{fig:hostingAI}
    \vspace{-4mm}
\end{figure}

\section{Hosting AI Workloads}
\label{sec:hosting}

\Cref{fig:hostingAI} shows the stack required to host AI workloads within a cloud provider:
from datacenter infrastructure and specialized hardware to the workloads.
To minimize the TCO of AI datacenters, we start at the top of this stack by analyzing AI workloads and reasoning about the demands they place on the underlying hardware and infrastructure.

\subsection{AI Workloads}
\label{sec:ai_workloads}
Nowadays, cloud providers host a wide range of AI workloads, spanning large language models (LLMs), vision and multimodal models, speech, recommendation systems, and classical deep neural networks (DNNs)~\cite{tpuv4,dxpu,metachip}. 
These workloads differ substantially in compute complexity, memory footprint, performance and accuracy targets, and input modalities.
The largest difference is between training and inference workloads:
training demands high-bandwidth memory, fast interconnects, and fault-tolerant checkpointing, while inference workloads range from latency-sensitive, memory-bound LLMs at small batch sizes to throughput-oriented vision and recommendation pipelines.

In this paper, we focus on LLM inference, which is rapidly becoming the dominant workload in AI datacenters~\cite{tpuv4,metachip,splitwise,polca}. 
For this workload, the most critical factors for datacenter build and provisioning are the size and architecture of the models (which drive compute and memory needs) and the user demand (the load that the system must sustain).

\myparagraph{AI Model Trends}
These workloads have rapidly evolved in scale, architecture, and demand over the past decade.

\begin{figure}[t]
    \centering
    \includegraphics[width=\linewidth]{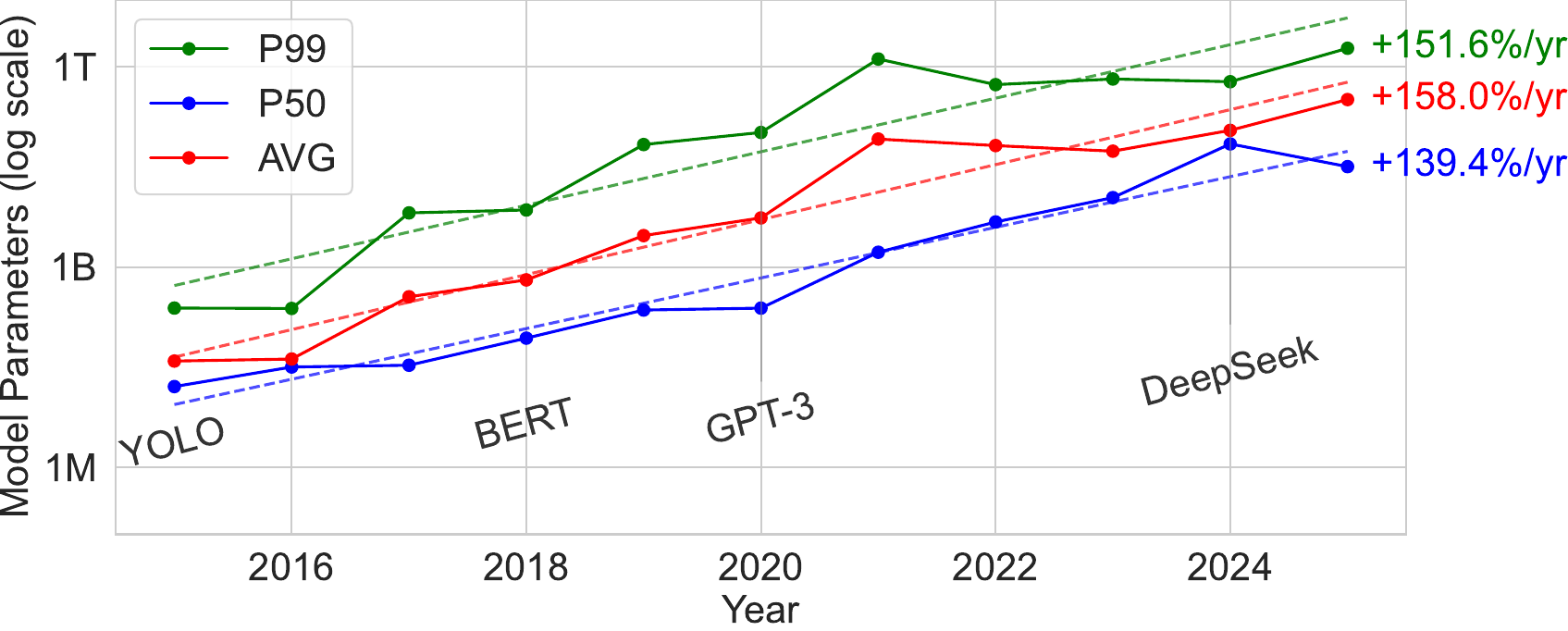}
    \vspace{-8mm}
    \caption{The P50, P99, and average size of the most popular AI models published in the last decade.
    }
    \label{fig:models_over_years}
    \vspace{-3mm}
\end{figure}

\myparagraphemph{Scale}
Models have grown dramatically in size, demanding far more compute, memory, and interconnect bandwidth.
Larger models drive higher FLOP requirements for inference and require larger on-device memory with greater bandwidth to store weights and intermediate results.
Once models exceed the capacity of a single GPU, multi-GPU setups become necessary, connected by high-bandwidth, low-latency links to exchange activations efficiently.
\Cref{fig:models_over_years} shows this trend: model sizes have increased exponentially over the past decade, from Google Neural Machine Translation’s $\approx$200M parameters in 2016~\cite{gnmt}, to GPT-3’s 175B parameters in 2020~\cite{gpt3}, and most recently to Llama 4 Behemoth, which exceeds 2T parameters~\cite{llama4behemoth}.

However, recent trends indicate a slowdown, with growth shifting to linear~\cite{epoch2024frontierlanguagemodelshavebecomemuchsmaller} and potentially becoming sublinear or even flat in the medium term.
This plateau reflects diminishing returns from traditional LLM scaling and reduced gains in training efficiency.
Several emerging approaches are attracting attention as alternatives to pure model scaling.
For example, distillation transfers knowledge into smaller, more efficient models~\cite{xu2024survey}; and
reasoning models leverage time-test-compute to enhance capabilities without relying on parameter growth~\cite{chen2025towards, guo2025deepseek}.

\myparagraphemph{Architecture}
Model architecture largely determines how much computation and memory bandwidth an inference workload needs.
Transformer-based models dominate current deployments~\cite{vaswani2017attention}. 
Their attention layers scale quadratically in compute and memory with sequence length, while large-kernel matrix multiplications (GEMM) demand high FLOP throughput and sustained memory bandwidth.
Emerging alternatives like state-space models (SSMs)~\cite{gu2022efficientlymodelinglongsequences} replace attention with convolution-like operations, reducing memory footprint and improving scalability for long contexts.

Mixture-of-Experts (MoE)~\cite{shazeer2017outrageouslylargeneuralnetworks} activate only a subset of experts per query, lowering the average compute cost but increasing the memory and network demands due to expert sharding.
Complementing these designs, fine-tuning techniques such as low-rank adaptation (LoRA)~\cite{hu2021lora} update only a small subset of parameters, reducing memory and bandwidth needs for model updates. 

Despite architectural differences, the core computation across modern AI models reduces to large matrix multiplications.
This commonality enables a unified performance modeling framework for AI inference—unlike traditional general-purpose datacenters, where heterogeneous workloads (\eg{}, databases, web services, and scientific applications) made unified modeling impractical.
Even as emerging applications adopt agentic systems that orchestrate multiple LLMs to solve complex, multi-step tasks~\cite{acharya2025agentic, chaudhry2025murakkab}, they still rely on foundational models for each component stage, preserving a consistent underlying computational structure. 

\myparagraph{User Demand} 
The global AI market is projected to grow from \$638B in 2024 to over \$3.68T by 2034~\cite{precedence2024}, with U.S. generative AI expected to see a 36.3\% CAGR through 2030~\cite{grandview2024}.
This growth is driving increased inference workloads, which already dominate AI operational costs~\cite{polca}. 
Unlike training, inference incurs higher cumulative costs due to continuous, large-scale deployment serving millions of queries daily~\cite{statChat}.
Cloud providers like Microsoft, Amazon, and Google report 15--25\% year-over-year growth in AI workloads~\cite{azure2025growth,aws2025growth,google2025growth}, reflecting rising user demand and the shift toward scalable, cost-efficient inference infrastructure.

\subsection{Hardware for AI Workloads}

\myparagraph{Accelerators}
AI inference workloads are both compute- and memory-intensive, relying heavily on GEMM and attention mechanisms.
These kernels require high floating-point throughput and high-bandwidth memory (HBM) to sustain performance.
Modern accelerators (\eg{}, GPUs~\cite{a100,h100,amdMI}, TPUs~\cite{tpuv4}, and NPUs~\cite{npu1,npu2}) combine massive parallelism with optimized memory systems that offer both large capacity 
and sufficient bandwidth to feed compute efficiently.

\myparagraph{Interconnects}
AI servers rely on high-speed interconnects to fully utilize accelerators. 
These links—electrical or optical, over copper, active optical cables, or co-packaged optics—differ in bandwidth, latency, and energy efficiency.
\emph{Intra-node} connections (PCIe, NVLink~\cite{nvlink}) enable fast device communication within a server, while \emph{inter-node} networks (InfiniBand, RoCE~\cite{roce}) are critical for scaling large models across servers.
Efficient collective operations (\eg{}, all-reduce, all-to-all) are essential to synchronize activations, gradients, and KV-cache data without stalling computation~\cite{nccl,gond2025tokenweaveefficientcomputecommunicationoverlap}. 

\subsection{Datacenter Infrastructure for AI Hardware}
The high performance of modern AI accelerators comes with high demands on supporting infrastructure.
Power, cooling, and networking requirements for large-scale AI deployments far exceed those of traditional datacenters. 

\myparagraph{Power}
High-end accelerators consume hundreds of watts per device.
For example, an NVIDIA DGX H100 server with eight GPUs has a thermal design power (TDP) of 10.2kW~\cite{h100}.
As a result, rack densities can range from several kilowatts to over 100 kW per rack~\cite{highdensity}.
Sustaining such loads requires robust \emph{power delivery} systems, including high-capacity uninterruptible power supplies (UPS) and power distribution unit (PDU) topologies, often with oversubscription to balance utilization and provisioning costs~\cite{flexDatacenter}. 
In addition, rapid workload fluctuations can induce large transient currents~\cite{li2024unseenaidisruptionspower,powerstabilization}, requiring careful electrical design and monitoring to maintain stability.

\myparagraph{Cooling}
The heat output of dense AI clusters quickly exceeds the practical limits of traditional air cooling~\cite{stojkovic2025tapas}. 
To maintain performance and reliability, operators increasingly adopt advanced \emph{cooling} solutions such as rear-door heat exchangers, direct-to-chip liquid cooling, and, in ultra-dense deployments, full liquid cooling~\cite{blackwellLiquid}. 
While these technologies enable sustained operation, they also require specialized facility layouts, coolant distribution systems, and higher maintenance complexity.

\myparagraph{Networking}
Large-scale AI inference also stresses \emph{networking} infrastructure. 
Models using tensor, pipeline, or expert parallelism rely on low-latency, high-bandwidth communication between thousands of accelerators~\cite{tpuv4}. 
This drives adoption of specialized network topologies (\eg{}, fat-tree, dragonfly) and high-performance interconnect technologies such as InfiniBand and RoCE~\cite{roce}, 
with precise bandwidth provisioning to prevent collective communication from becoming the bottleneck. 
The capital cost of network switches, optical modules, and cabling for such deployments can represent a significant share of overall system expenditure.

In combination, these infrastructure requirements make AI datacenter builds substantially more complex and capital-intensive than traditional deployments.

\section{Datacenter Lifecycle}

Based on these AI workloads, hardware, and infrastructure trends, we examine the datacenter lifecycle to identify opportunities for reducing TCO.
\Cref{table:stages} breaks the lifecycle into: \emph{build}, \emph{IT provisioning}, and \emph{operate}.
These stages help us explore how traditional lifecycle policies must be revisited to address the scale, density, and performance demands of AI.
On this foundation, we develop a TCO 
model to evaluate costs across design choices over the datacenter’s lifetime.

\begin{table}[t!]
\vspace{2mm}
\centering
\footnotesize
\begin{tabular}{lll}
\toprule
\textbf{Stage} & \textbf{Description} & \textbf{Timeline} \\
\midrule
\emph{Build} & Site selection and facility construction & 15--30 years\\
\emph{IT provision} & IT hardware deployment and upgrades &  4--6 years\\
\emph{Operate} & Workload scheduling, resource management &  Per inference \\
\bottomrule
\end{tabular}
\caption{Lifecycle stages for datacenter infrastructure.}
\label{table:stages}
\vspace{-7mm}
\end{table}

\subsection{Stages of a Datacenter Lifecycle}

\myparagraphemph{Build}
This stage is where the datacenter facility itself is designed and constructed. 
Decisions made here set long-lived constraints on 
utility capacity, 
substation feeds, 
power distribution topology (\eg{}, flat \vs{} hierarchical),
cooling technology (air \vs{} liquid), floor space, and network fabric. 
These choices determine the maximum achievable rack density, 
define fault domains, and 
affect how easily the facility can accommodate future upgrades such as liquid cooling or higher-voltage power buses. 
Networking decisions (\eg{}, Ethernet \vs{} InfiniBand, optical link reach, and oversubscription ratios) impact job scaling efficiency and the cost of east--west traffic, which is critical for large AI models distributed across accelerators.

\myparagraphemph{IT provisioning}
This stage defines when and how to introduce new accelerators and retire or repurpose older ones. 
These decisions balance several factors, including performance-per-watt improvements from newer hardware, 
cost of newer hardware, software maturity and compatibility, 
depreciation schedules, and risk of underutilized power or cooling.
IT provisioning may involve mixed-generation GPU pools with placement constraints or repurposing older GPUs to give them a ``second life'' on workloads with lower performance requirements (\eg{}, fine-tuning or batch analytics).

\myparagraphemph{Operate}
In this stage, decisions focus on where to place model instances, 
how to schedule queries to instances, and 
how to execute queries efficiently. 
Placement strategies account for hardware heterogeneity, ensuring that each model runs on the accelerator generation that provides the best performance-to-cost tradeoff.
Scheduling considers factors such as service-level objectives (SLOs), 
model routing based on query complexity,  
and price-aware policies that shift flexible workloads across time or geography. 
Execution leverages AI-specific optimizations, including dynamic batching, quantization, speculative decoding, distillation, and model disaggregation,
to reduce cost per query while meeting SLOs.

\myparagraph{Traditional Approach}
\Cref{table:traditionalDCLifecycle} outlines the lifecycle of general-purpose datacenters, which are primarily powered by CPUs and designed to support diverse workloads. 

\begin{table}[t]
\vspace{0mm}
\centering
\footnotesize
\begin{tabular}{p{13mm} p{65mm}}
\toprule
\textbf{Stage} & \textbf{Traditional Approach Characteristics} \\
\midrule
\emph{Build} & Hierarchical power distrib; Air cooling; Ethernet network. \\
\emph{IT provision} & Fixed per-server lifecycle; New server generations released every 2--3 years; Gradual replacement. \\
\emph{Operate} & Services tied to fixed hardware configurations; Instances migrated to new hardware when released; Legacy applications remain on old servers. \\
\bottomrule
\end{tabular}
\caption{Overview of the traditional approach to manage the lifecycle of a traditional datacenter for general-purpose serving CPU-based workloads.}
\label{table:traditionalDCLifecycle}
\vspace{-6mm}
\end{table}

\myparagraphemph{Build}
Traditional datacenters rely on a conservative, uniform infrastructure. 
Power distribution usually follows a hierarchical topology:
from the colo-level to rows, and then to individual racks, with each level having its own power caps~\cite{flexDatacenter}. 
Cooling is primarily air-based, and networking is implemented with standard Ethernet~\cite{bilal2013quantitative,greenberg2008cost}.

\myparagraphemph{IT provisioning}
Servers follow a fixed lifecycle, with each generation operating for a defined period before decommissioning.
New hardware is typically released every 2--3 years, and operators replace older servers in a phased manner according to this schedule.

\myparagraphemph{Operate}
Services run on servers with a specific hardware configuration.
When new hardware is introduced, new services migrate to the latest generation, while legacy applications remain on older servers.
This approach prioritizes stability and predictability but reduces flexibility to exploit hardware heterogeneity or tailor performance to specific workloads. 

\begin{table*}[t]
\centering
\footnotesize
\begin{tabular}{p{10mm} p{20mm} p{100mm} p{28mm}}
\toprule
\textbf{Category} & \textbf{Component} & \textbf{Description} & \textbf{Example Cost (\$)} \\
\midrule
\multirow{5}{*}{\emph{CapEx}} 
& IT & Servers, racks, accelerators, storage & \$375k/server~\cite{h100Cost} \\
& Networking & Fabric switches, optics, structured cabling & \$2000/server~\cite{nvlink,networkopcost1} \\
& Building & Site preparation, building shell, land, electrical and mechanical base infrastructure & \$0.5/$\text{ft}^2$~\cite{inigointelligentplacement} \\
& Power & Power infrastructure (Switchgear, transformers, UPS, PDUs, busbars, rack distribution) & \$7.0/W~\cite{powerBarroso,howmuchcost} \\
& Cooling & Cooling infrastructure (chillers, CRAH/CRAC units, pumps, piping, liquid loops, airflow) & \$2.5/W~\cite{howmuchcost,liquidvsair} \\
\midrule
\multirow{4}{*}{\emph{OpEx}} 
& Networking & Port licenses, optics replacement, networking component power & \$600/server~\cite{networkopcost1,networkopcost2,networkopcost3} \\
& Energy & IT load scaled by PUE, utility tariffs, demand charges & \$20--40/MWh~\cite{gridstatus2025} \\ 
& Maintenance & Spares, repairs, monitoring, water/treatment, field-replaceable units, failure-rate & \$5000/server~\cite{maintenancecost} \\ 
& Software & Licenses, support contracts & \$200/server \\ 
\bottomrule
\end{tabular}
\caption{TCO components for an example datacenter with DGX H100~\cite{h100Cost} servers.
}
\label{table:tco_components}
\vspace{-7mm}
\end{table*}

\myparagraph{Rearchitecting for AI}
As shown in \Cref{sec:hosting}, AI workloads challenge traditional datacenter design.
Modern accelerators have higher power and thermal demands, increasing the value of liquid cooling and high-density racks.
Space and density constraints needs favor scale-up architectures like NVLink Switch-based designs~\cite{nvlink}.

\emph{Memory capacity and bandwidth} have grown to support larger model contexts, enabling more complex workloads but raising per-node costs.
Efficient memory provisioning is essential.
Similarly, \emph{interconnect} performance across servers is critical for parallel efficiency (low-bandwidth or high-latency links can severely limit scaling).

Trade-offs between cost and performance 
must be evaluated both within each stage (build, IT provisioning, and operate) and across them.
For example, separating prefill and decode across servers~\cite{splitwise} enables using heterogeneous hardware and influences \emph{build} and \emph{refresh} strategies.

\subsection{Total Cost of Ownership (TCO) Model}
\label{sec:tcomodel}

We use a \emph{comprehensive TCO model} that enables automated, end-to-end lifecycle analysis by capturing workload dynamics, hardware evolution, and system aging.
It accounts for costs across lifecycle stages and supports exploration of how workload trends, hardware roadmaps, and infrastructure decisions impact total cost.

\myparagraph{Model overview}  
\Cref{table:tco_components} summarizes the components of the TCO, breaking them down into \emph{CapEx} and \emph{OpEx}:

\noindent
\begin{flushleft}
$\mathrm{TCO} = \mathrm{CapEx} + \mathrm{OpEx}$
\end{flushleft} 
\noindent
\begin{flushleft}
$\mathrm{CapEx} = \mathrm{CapEx}_{\mathrm{Facility}} + 
\mathrm{CapEx}_{\mathrm{Power}} + 
\mathrm{CapEx}_{\mathrm{Cool}} + 
\mathrm{CapEx}_{\mathrm{Net}} + 
\mathrm{CapEx}_{\mathrm{IT}}$
\end{flushleft}
\noindent
\begin{flushleft}
$\mathrm{OpEx} = \mathrm{OpEx}_{\mathrm{energy}} +\mathrm{OpEx}_{\mathrm{M\&R}} +  \mathrm{OpEx}_{\mathrm{network}} + \mathrm{OpEx}_{\mathrm{software}}$
\end{flushleft} 

For the annualized TCO, 
$\mathrm{CapEx}$ amortizes long-term infrastructure and IT over their useful lives, $\mathrm{OpEx}$ captures variable and recurring operational costs over a full year. 

\begin{figure}[t]
    \centering
    \includegraphics[width=\linewidth]{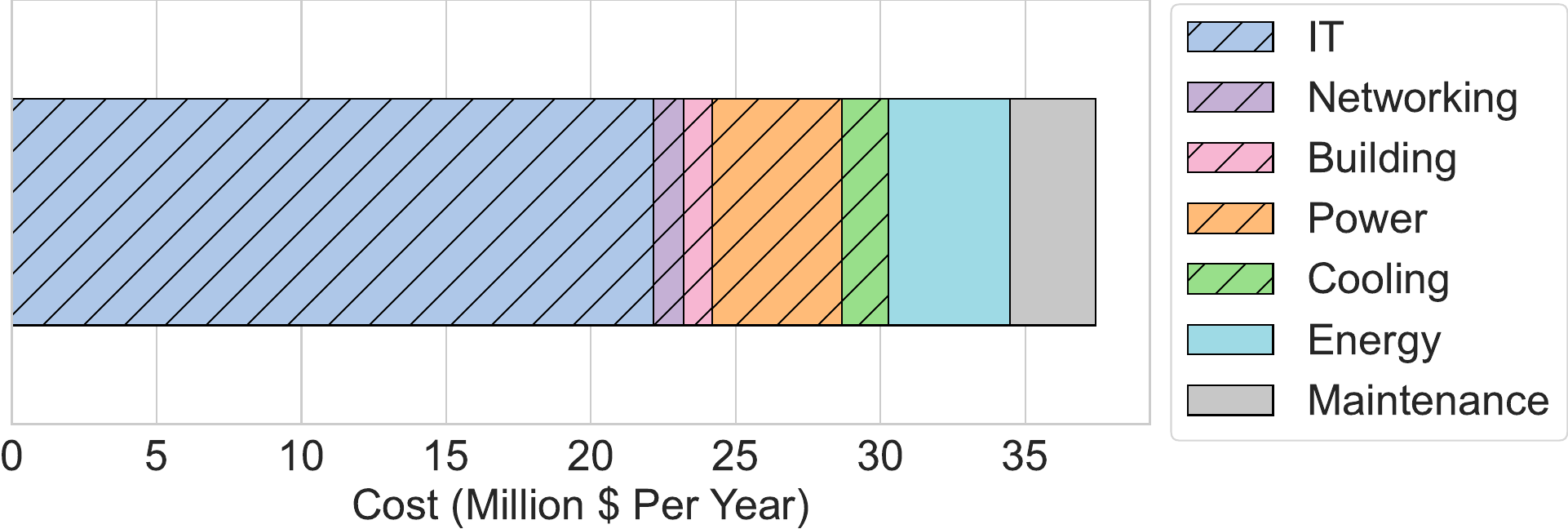}
    \vspace{-8mm}
    \caption{TCO breakdown for a 10MW AI datacenter.
    }
    \label{fig:tco_example}
    \vspace{-5mm}
\end{figure}

\Cref{fig:tco_example} shows the annual datacenter TCO breakdown, divided into CapEx and OpEx categories for a representative user demand and model size projected for 2025.  
At 75\% average utilization, this 10MW datacenter~\cite{flexDatacenter} sustains roughly 500 H100 servers, consuming 70 GWh of energy per year for operation.
GPU servers drive IT CapEx and dominate costs, followed by energy-related OpEx.
On the other hand, building construction and maintenance contribute the least.

\myparagraph{Capital Expenses (CapEx)}
This is the upfront costs for acquiring, building, or upgrading long-term datacenter assets, 
including facility, IT equipment (such as servers and racks), and networking infrastructure.
Facility, network, and IT assets are amortized over 15--30, 7--10, and 3--5 years respectively, using straight-line or declining-balance depreciation. 
CapEx is often normalized per delivered kW or per accelerator to facilitate design comparisons.

\myparagraphemph{IT Infrastructure}
It consists of compute servers equipped with CPUs and accelerators such as GPUs, TPUs, or NPUs, along with racks and storage devices like NVMe. 
Modern servers may also include CXL-attached memory to expand capacity. 
These components provide the core computational and storage resources of the datacenter.

\myparagraphemph{Networking Infrastructure}
It connects servers, storage, and other datacenter resources. 
Fabric switches route traffic between racks, while optical transceivers and structured cabling provide high-bandwidth, low-latency connections across the facility.
This infrastructure ensures efficient communication for distributed workloads.

\myparagraphemph{Building}
It includes the physical infrastructure required to support a datacenter.
These foundational elements ensure a safe and efficient environment for all equipment.

\myparagraphemph{Power Infrastructure}
Electrical systems deliver reliable power to racks and IT equipment, including all power devices and connections to ensure stable and redundant power distribution throughout the datacenter. 

\myparagraphemph{Cooling Infrastructure}
Mechanical systems cool datacenters to maintain safe operating temperatures, with core infrastructure such as chillers and pumps. 
Together, these systems prevent overheating and enable high-performance operation.

\myparagraph{Operational Expenses (OpEx)}
This is the ongoing costs of running and maintaining a datacenter.
We model utilization-sensitive costs (\eg{}, energy) based on workload mix and scheduling policies, reflecting how activity levels impact overall consumption.
In contrast, utilization-insensitive costs (\eg{}, fixed maintenance, software contracts, and leases) are treated as per-rack or per-site constants, providing a stable baseline that does not fluctuate with workload intensity.

\myparagraphemph{Network Operations}
Covers the ongoing costs of maintaining the datacenter network, including licensing fees for switch ports and the replacement of failed components. 
Costs are influenced by network size, redundancy, and utilization patterns, as higher traffic and denser topologies can increase wear and require more frequent upgrades or maintenance. 

\myparagraphemph{Peak power}
Datacenters often pay monthly fees based on their highest instantaneous power draw.
Utilities charge this way because the grid must be provisioned for these peaks: generators, transformers, and transmission lines all need capacity for the maximum load, even if brief.
Peaks can also threaten grid stability and require fast-response resources like spinning reserves.
By tying costs to peak usage, utilities incentivize datacenters to smooth load spikes, easing stress on the grid and lowering overall infrastructure costs.

\myparagraphemph{Energy}
They include the electricity consumed by IT equipment as well as the supporting infrastructure (\eg{}, cooling and power distribution systems).
These costs, typically billed on a monthly basis, account for datacenter's IT hardware utilization and \emph{power usage effectiveness} (PUE), which measures the ratio of total facility energy to energy used by IT equipment.
A higher PUE indicates more energy spent on overheads such as cooling and power conversion.

\myparagraphemph{Maintenance \& Repairs}
Includes both preventive and corrective maintenance of mechanical, electrical, and IT systems. Costs are primarily driven by the expected failure rates of components such as cooling and power components, servers, storage devices, and networking equipment. This category also covers other maintenance ensuring that the datacenter remains operational and reliable over time.

\myparagraphemph{Other}
Recurring expenses such as software licenses, support contracts, and land or lease payments. 
These costs are largely fixed and do not vary with workload or utilization.

\section{TCO-Driven Lifecycle Framework}
\label{sec:methodology}

To evaluate the long-term economics of AI datacenters, we develop \emph{a cross-stage, TCO-driven framework} that spans the 15-year datacenter lifecycle, covering
\emph{build}, \emph{IT provisioning}, and \emph{operation}.
Our framework couples workload dynamics, model evolution, hardware road maps, and infrastructure costs to project scenarios and assess alternative policies, allowing us to identify the best policies both within each stage and across stages.
We validate our methodology against real-world observations.

\subsection{Modeling Assumptions}

\myparagraph{Timeline}
We model a lifecycle of 15 years starting from 2015 and ending in 2030.
The methodology generalizes to longer or shorter horizons, but uncertainty increases the further we project.
For validation, we fit model outputs against current-day trends (\ie{}, 2025), then forecast costs and fleet composition for the following five years.

\myparagraph{Workload}
We focus on LLM inference as the dominant AI datacenter workload~\cite{foundation2024inference}.
Training workloads would follow a similar methodology but differ in modeling, as they are heavier in both computation and communication.
We use input traces from DynamoLLM~\cite{stojkovic2024dynamollm}, which exhibit diurnal patterns, and assume a baseline of 100K requests per second.
Following \Cref{sec:ai_workloads}, we apply a 15\% annual growth rate~\cite{azure2025growth,aws2025growth,google2025growth}, which implies over 200K RPS after five years.

\myparagraph{AI Models}
Based on 2015--2025 parameter scaling trends (\Cref{sec:ai_workloads}), we assume linear growth in model size through 2030, with alternative scenarios for accelerated (exponential) or slowed (sub-linear) scaling.
Providers are assumed to adopt new models via \emph{smooth migration}, gradually transitioning workloads rather than abrupt cutovers~\cite{distilation1,distilation2}.
We also assume future LLMs follow the LLaMA design lineage~\cite{llama}, \ie{}, decoder-only transformer architectures with consistent layer organization, attention mechanisms, and parameterization strategies.
This assumption reflects the convergence of recent frontier models, which primarily differ in scale.

\myparagraph{Hardware}
Hardware projections include FLOPS, memory bandwidth, TDP, and cost, with linear growth trends~\cite{epoch2023trendsinmachinelearninghardware}.
We also model delays between the announcement of a new GPU~\cite{nvidia_b200_release} and its actual mass availability in cloud providers~\cite{aws_b200_release,gc_b200_release,azure_b200_release} (\eg{}, B200 had a delay between 6 months and a year). 

\myparagraph{Performance}
We develop a roofline model tailored to LLM inference across diverse hardware configurations.
Our validation against known model==hardware pairings shows that it aligns with prior work~\cite{yuan2024llminferenceunveiledsurvey,guo2025systemperformancecostmodelinglarge} and with profiling results from real workloads.
The model captures how hardware limits (compute throughput and memory bandwidth) interact with workload characteristics (arithmetic intensity and memory footprint) during LLM inference).

Our model derives the arithmetic intensity and memory footprint analytically from the LLM architecture and parameter count, making the predictions reproducible, validated against profiling-based approaches.
Extending the model to new GPUs requires only the peak FLOPs and memory bandwidth (GB/s) of the device, while applying it to new LLMs requires recomputing theoretical compute and memory requirements from the model’s structure.
The roofline model predicts the time-to-first-token (TTFT) and time-between-tokens (TBT) latencies for a given hardware, model, and request load (\eg{}, H200 running Llama3 at 10 requests per second will have a TTFT of 200 ms and a TBT of 50 ms).

We then increase the load until requests exceed an SLO of 400 ms for TTFT and 100 ms for TBT~\cite{stojkovic2024dynamollm}.
The resulting goodput is the maximum request rate (RPS) the system can sustain without violating the SLO. 
This approach identifies, for each hardware–model pair, the utilization point at which latency begins to degrade.
Using this SLO, we provision the minimal number of GPUs required to serve the load and calculate the corresponding GPU utilization.

\subsection{Optimization Goal}

\myparagraph{Cost}
The framework integrates both CapEx (IT hardware, networking, building, power, cooling) and OpEx (networking, energy, maintenance, software).
CapEx is amortized over useful lifetimes (\eg{}, 15--30 years for facilities, 3--5 years for GPUs), while OpEx captures recurring operational costs.
This enables evaluation of trade-offs such as upfront investment in cooling vs. deferred savings in refresh.

\myparagraph{Carbon}
A similar methodology applies to carbon emissions, including embodied carbon from construction and hardware, and operational carbon from electricity use.
We leave this extension to future work.

\subsection{Lifecycle Evaluation}

\begin{figure}[t]
    \includegraphics[width=\columnwidth]{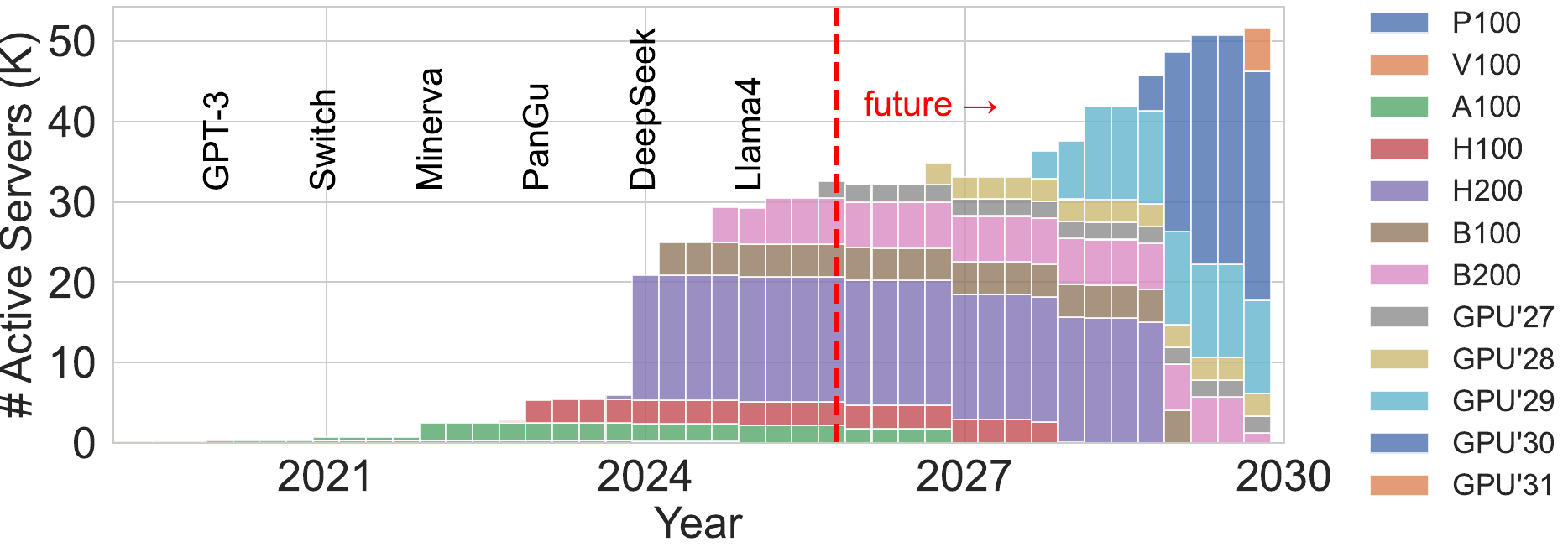}
    \vspace{-7mm}
    \caption{Server count by GPU type over time in an AI fleet following the traditional baseline in \Cref{table:traditionalDCLifecycle}.
    Includes the release dates of major AI models. 
    }
    \label{fig:server_count_baseline}
    \vspace{-3mm}
\end{figure}

\myparagraph{Baseline Timeline}
\Cref{fig:server_count_baseline} shows the simulated deployment timeline of an AI fleet under the baseline policy, which follows the traditional datacenter lifecycle approach (\Cref{table:traditionalDCLifecycle}).
The figure shows how model release cycles and hardware availability shape the fleet composition over time, with the release dates of notable large models marked for reference.
The simulation begins in 2015 with 50 P100 GPU servers sustaining an initial steady-state load of 100K RPS.
At this point, the total annual TCO of the AI datacenter is $\approx$\$0.2M.

As user demand and model size grow (\ie{}, 15\% year-over-year), the fleet scales gradually.
By 2024, traffic reaches 350K RPS, coinciding with the release of DeepSeek V3, a 671B-parameter model~\cite{deepseekv3,liu2024deepseek}, which triggers a major hardware refresh with H200 GPUs.
This refresh increases the total server count to 25K to meet performance targets, and the annual TCO rises to $\approx$\$0.3B, reflecting the added hardware, expanded datacenter infrastructure, and higher operating expenses.
Peaks in server deployments align with major LLM launches, highlighting the close connection between AI model roadmaps and datacenter economics.

\myparagraph{Solution Approach}
We run Monte Carlo simulations~\cite{metropolis1949monte} to capture uncertainty in workload growth, model scaling, hardware availability, and cost projections.
Each simulation samples these distributions, yielding a range of outcomes for TCO under different policies (\eg{}, aggressive vs. delayed server refresh).
We use the traditional approach in \Cref{table:traditionalDCLifecycle} as a baseline.
By comparing distributions, we identify which decisions dominate long-term cost and where flexibility provides the greatest value. 

\section{Building Efficient AI Infrastructure}
\label{sec:stage_build}

The first stage in the datacenter lifecycle is building the infrastructure, which includes:
(1) the physical building that houses the datacenter,
(2) the power delivery that supplies electricity to servers and other equipment,
(3) the cooling that removes the heat generated by servers, and
(4) the networking that interconnects servers.

The physical building is relatively static, while power, cooling, and networking requirements~\cite{powerBarroso} must anticipate shifts in workload demand and hardware capabilities over decades of operation.
Emerging AI workloads, with their extreme power and thermal densities and distinct communication patterns, are already reshaping the design space across these dimensions.
We revisit these design choices using our TCO-driven framework to build a holistic understanding over the full lifecycle and validate our approach comparing it to the directions already being pursued in practice.

\subsection{Power Infrastructure}

\begin{figure}[t]
    \centering
    \includegraphics[width=\linewidth]{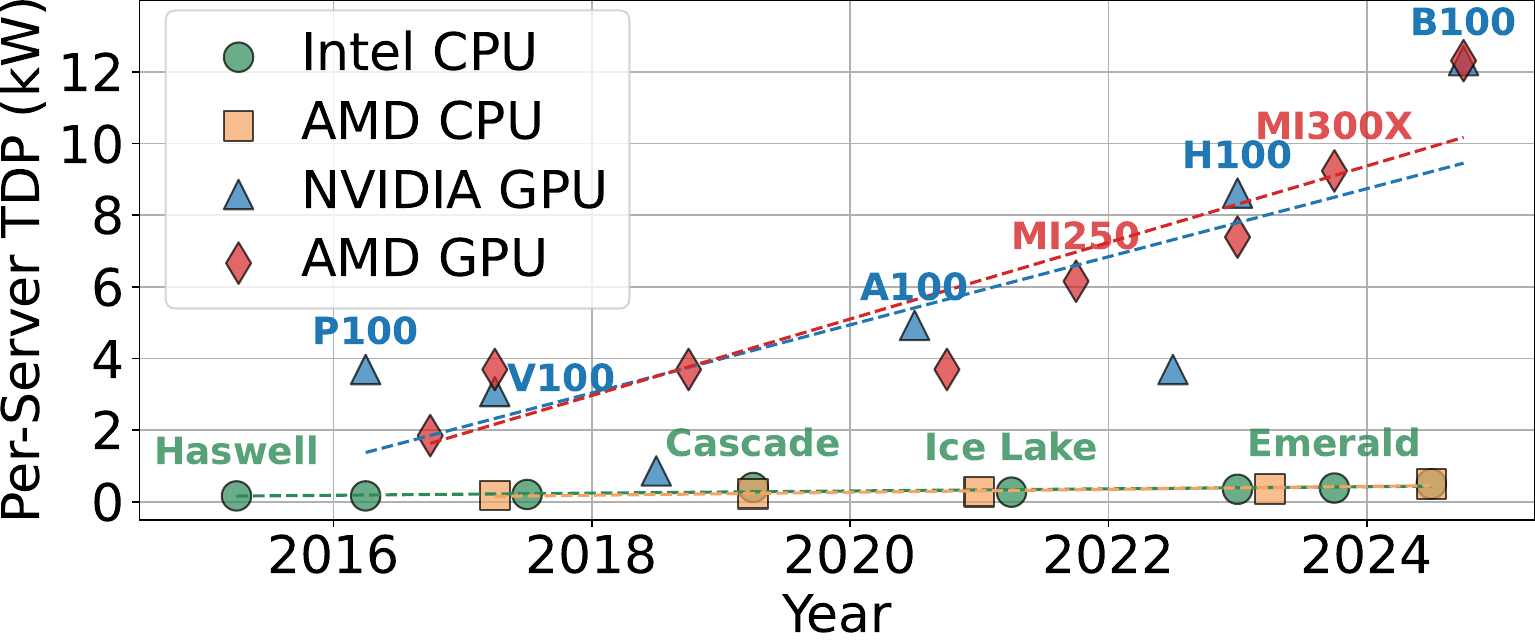}
    \vspace{-7mm}
    \caption{Per-server TDP across Intel and AMD CPUs, and NVIDIA and AMD GPUs over the years.}
    \label{fig:tdp_gpu_cpu}
    \vspace{0mm}
\end{figure}

\myparagraph{Traditional Approach}
Traditional datacenters adopt a hierarchical power distribution, balancing fault isolation, maintainability, and power stranding~\cite{wang2009ship, flexDatacenter, wu2016dynamo}.
An Automatic Transmission Switch (ATS) directs power from the grid to multiple Uninterruptible Power Supplies (UPS) with redundancy.
Each UPS supports a fraction of the total load and feeds downstream into multiples Power Distribution Units (PDUs), distributing power to server rows and racks. 

Server and rack provisioning must respect capacity limits within each power sharing domain defined by the hierarchy. When the budget at a power domain (\eg{}, per-PDU) is $X$ and each server consumes $Y$, only $\lfloor X / Y \rfloor$ servers can be deployed.
The residual capacity, $X - Y \times \lfloor X / Y \rfloor$, becomes stranded power (\ie{}, power fragmentation). 

\begin{table}[t!]
\footnotesize
\centering
\begin{tabular}{>{\bfseries}p{0.6cm} p{1.7cm} p{1.1cm} p{1cm} p{1.1cm}}
\toprule
 &         & \multicolumn{3}{c}{\textbf{Power domain}} \\
 & Feature & Per-PDU & Per-UPS & Per-DC \\
\midrule
\multirow{2}{*}{CapEx} 
& Stranding & \cellcolor{red!20}Lower  & \cellcolor{yellow!20}Medium & \cellcolor{green!20}Higher \\
& Complexity       & \cellcolor{red!20}Higher & \cellcolor{yellow!20}Medium & \cellcolor{green!20}Lower \\
\midrule
\multirow{1}{*}{OpEx}  
& Maintenance      & \cellcolor{red!20}Higher & \cellcolor{yellow!20}Medium & \cellcolor{green!20}Lower \\
\midrule
\multirow{1}{*}{Other} 
& Fault isolation  & \cellcolor{green!20}Excellent & \cellcolor{yellow!20}Good & \cellcolor{red!20}Poor \\
\bottomrule
\end{tabular}
\caption{Comparison of power delivery infrastructure designs.
Green: good, yellow: moderate, red: poor.
}
\label{tab:power_cost_comparison}
\vspace{-6mm}
\end{table}

\begin{figure*}[t]
  \centering
  \subfloat[TCO \vs{} power infrastructure.]{%
  \includegraphics[clip,width=0.57\columnwidth]{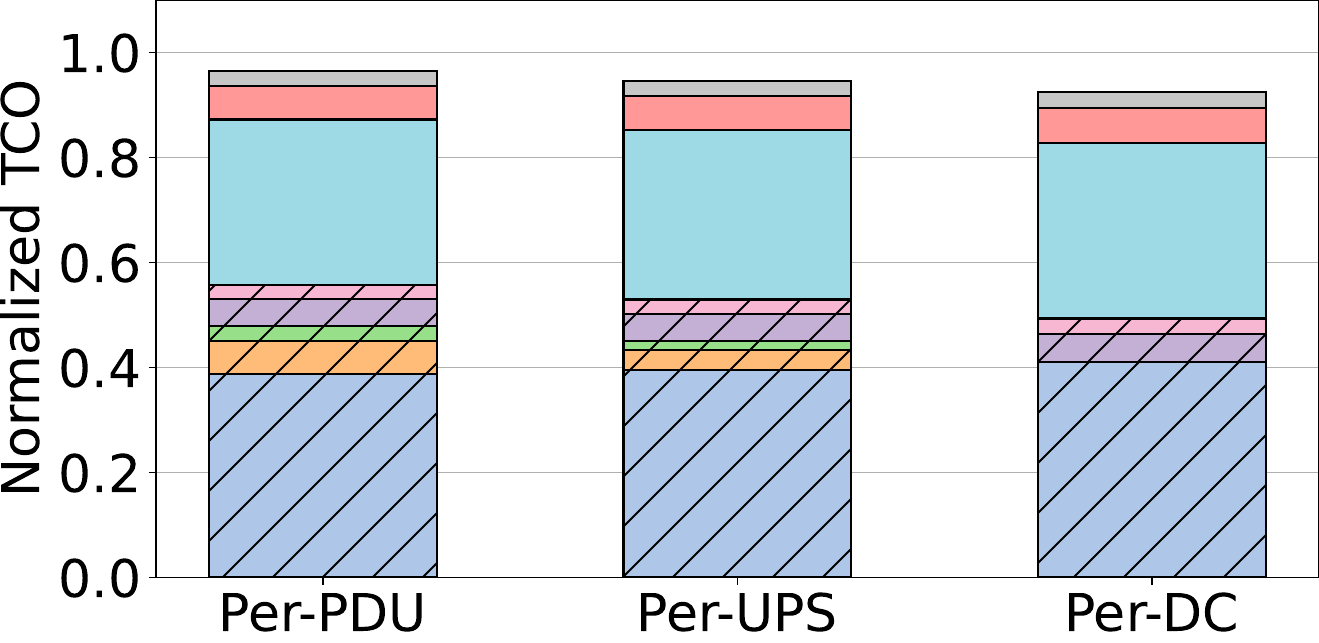}
  \label{fig:tco_power_infra}
  }
  \subfloat[TCO \vs{} cooling infrastructure.
  ]{%
  \includegraphics[clip,width=0.57\columnwidth]{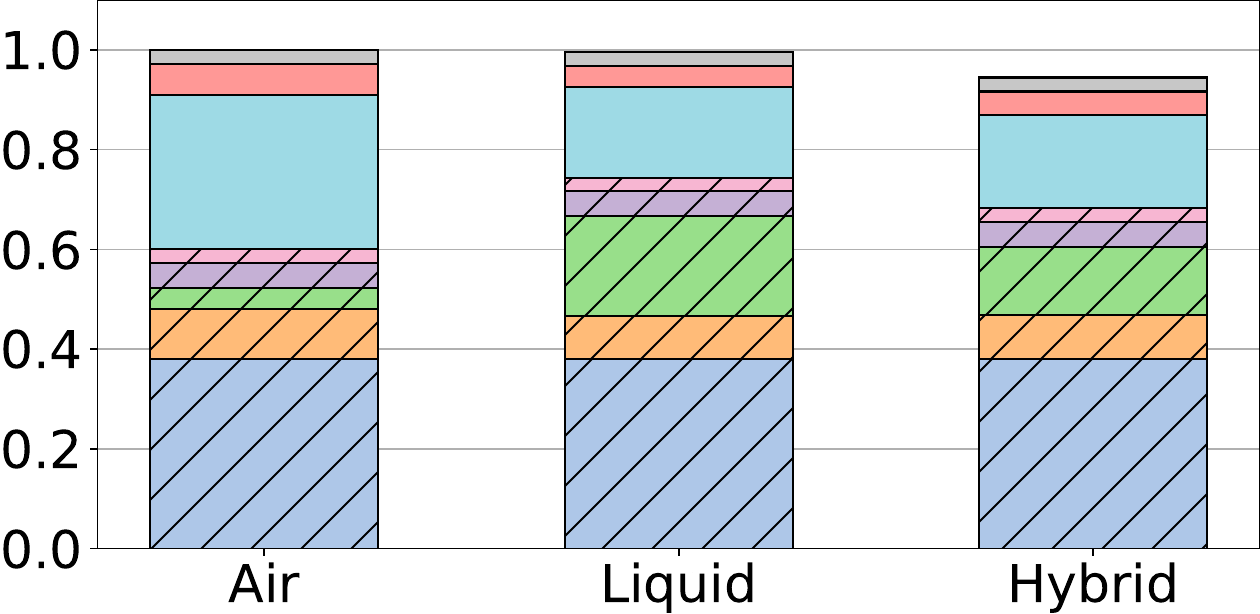}
    \label{fig:tco_cooling_infra}}
  \subfloat[TCO \vs{} networking infrastructure.
  ]{%
  \includegraphics[clip,width=0.93\columnwidth]{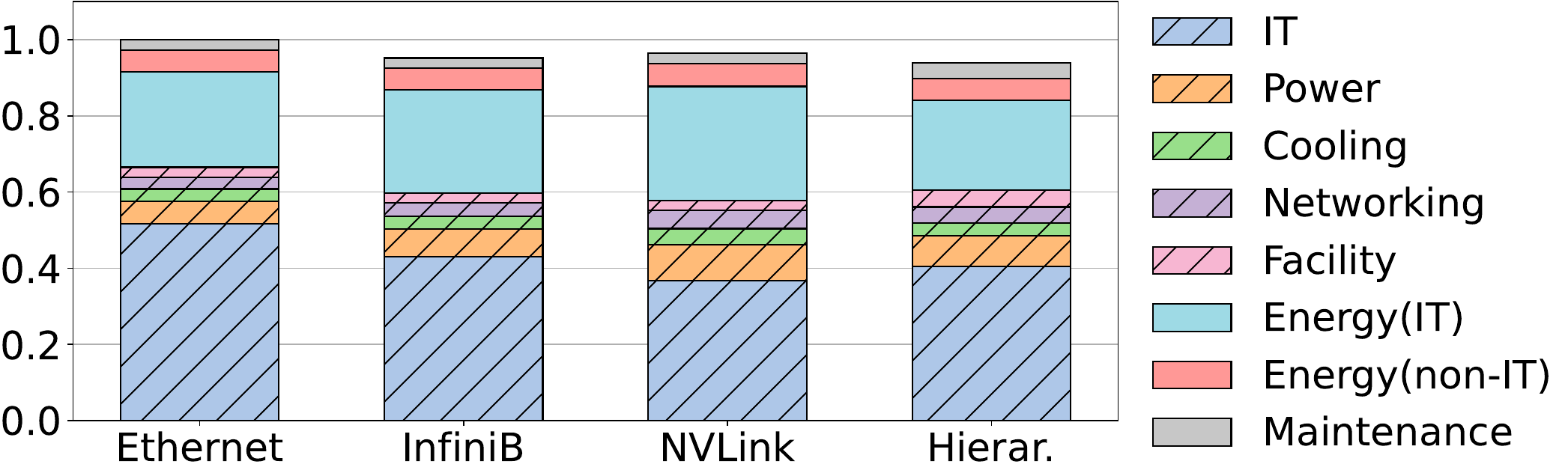}
    \label{fig:tco_networking_infra}}
        \vspace{-3mm}
 \caption{TCO \vs{} infrastructure designs during the \emph{build} stage.}
  \vspace{-3mm}
\end{figure*}

\myparagraph{Rearchitecting for AI}
AI accelerators (\eg{}, GPUs and TPUs) push power density far beyond historical norms.
\Cref{fig:tdp_gpu_cpu} shows that the TDP of high-end GPU servers has significantly outgrown the power demand of high-end CPU servers.
For example, while NVIDA DGX H100 server with 8 GPUs needs 10.2kW~\cite{h100}, Intel Emerald Rapids server with 64 cores needs only 385W~\cite{emrCost}. 

While this surge exacerbates power fragmentation, flatter power distribution architectures can pool power across broader domains to reduce stranding~\cite{google-mvpp, bianchini2024datacenter}.
However, these designs involve complicated trade-offs in fault tolerance, design and maintenance complexity, and TCO, we introduce simplifications by removing second-order effects and practical constraints (\eg{} supply chains) and summarize these trade-offs qualitatively in
\Cref{tab:power_cost_comparison}. 
For example, while a flat per-DC power delivery minimizes stranded capacity, it demands greater redundancy and more intricate maintenance planning.
Though a flat per-DC power delivery architecture might not be feasible in practice today, our TCO-driven analysis in \Cref{fig:tco_power_infra} shows that, with simplified modeling of associated costs and hardware requirements, it reduces lifecycle TCO by 4.2\% compared to conventional hierarchical designs~\cite{flexDatacenter}, motivating new designs enabling a flatter power sharing domains in real-world distribution schemes.

\subsection{Cooling Infrastructure}

\begin{table}[t]
\footnotesize
\centering
\begin{tabular}{>{\bfseries}p{0.6cm} p{2.0cm} p{1.4cm} p{1.4cm} p{1.4cm}}
\toprule
 & Feature & Air & Hybrid & Liquid \\
\midrule
\multirow{1}{*}{CapEx} 
  & Complexity & \cellcolor{green!20}Lower & \cellcolor{yellow!20}Medium & \cellcolor{red!20}Higher \\
\midrule
\multirow{2}{*}{OpEx} 
  & Energy efficiency & \cellcolor{red!20}Lower & \cellcolor{yellow!20}Medium & \cellcolor{green!20}Higher \\
  & Maintenance       & \cellcolor{green!20}Lower & \cellcolor{yellow!20}Medium & \cellcolor{red!20}Higher \\
\midrule
\multirow{2}{*}{Other} 
  & High-dense racks & \cellcolor{red!20}Lower & \cellcolor{yellow!20}Medium & \cellcolor{green!20}Higher \\
  & Noise level      & \cellcolor{red!20}Higher & \cellcolor{yellow!20}Medium & \cellcolor{green!20}Lower \\
\bottomrule
\end{tabular}
\caption{Comparison of cooling infrastructure designs.
}
\label{tab:cooling_comparison}
\vspace{-8mm}
\end{table}

\myparagraph{Traditional Approach}
Most datacenters rely on air-based cooling systems~\cite{meta-sustainability, energygov-evaporative, daraghmeh2017review, google-cooling, stojkovic2025tapas}, where chillers or adiabatic cooling units deliver cold air through air handling units (AHUs).
These AHUs circulate the air via raised floors or hot/cold aisle containment using fans.
Cooling capacity is provisioned conservatively to meet peak thermal loads.
Power usage effectiveness (PUE) is optimized through careful airflow management and economization. 
Modern datacenters achieve typical PUE values between 1.1 and 1.3~\cite{googlepue,microsoftpue,awspue}.

\myparagraph{Rearchitecting for AI}
High-density GPU racks generate far more heat than CPU-based system (often 4--8\texttimes{} higher per rack~\cite{blackwellLiquid}).
Meeting these loads requires significantly higher airflow rates, lower inlet temperatures, and more fan energy, pushing air cooling beyond its physical and economic limits.

As a result, liquid cooling (\eg{}, cold plates or immersion~\cite{overclockImmersion, google_liquid_cool}) is increasingly considered for future dense GPU deployments~\cite{blackwellLiquid, bianchini2024datacenter}. 
While upfront complexity and CapEx are higher, OpEx is reduced through improved heat transfer, reduced chiller load, and lower fan power, summarized in \Cref{tab:cooling_comparison}.
Hybrid designs, combining liquid for AI racks and air for low-density racks, balance cost, density, and maintainability.
Specifically, a hybrid design could be a mix of 75\% cold plates and 25\% air cooling~\cite{hybridcooling}, corresponding to the high and low density loads with a resulting Power Usage Effectiveness (PUE) of 1.15.
Our evaluation in \Cref{fig:tco_cooling_infra} shows that hybrid cooling  yields the lowest TCO with 9\% reduction over the full lifecycle.

\subsection{Networking Infrastructure}

\myparagraph{Traditional Approach}
General-purpose datacenters use multi-tier Ethernet fabrics (\eg{}, leaf--spine) with moderate oversubscription at higher tiers~\cite{bilal2013quantitative,greenberg2008cost}. 
This design is cost-effective for CPU-based workloads with modest inter-node bandwidth and latency needs. 

\begin{table}[t]
\footnotesize
\centering
\begin{tabular}{>{\bfseries}p{0.6cm} p{1.2cm} p{1cm} p{1.2cm} p{1cm} p{1.3cm}}
\toprule
 & Feature & Ethernet & InfiniBand & NVLink & Hierarchical \\
\midrule
\multirow{1}{*}{CapEx} 
  & Cost & \cellcolor{green!20}Lower & \cellcolor{yellow!20}Medium & \cellcolor{red!20}Higher & \cellcolor{yellow!20}Medium \\
\midrule
\multirow{2}{*}{OpEx} 
  & Energy & \cellcolor{green!20}Lower & \cellcolor{yellow!20}Medium & \cellcolor{red!20}Higher & \cellcolor{green!20}Higher \\
  & Maintain      & \cellcolor{green!20}Lower & \cellcolor{yellow!20}Medium & \cellcolor{red!20}Higher & \cellcolor{yellow!20}Medium \\
\midrule
\multirow{2}{*}{Perf} 
  & Bandwidth & \cellcolor{red!20}Lower & \cellcolor{green!20}Higher & \cellcolor{green!20}Higher & \cellcolor{green!20}Higher \\
  & Latency   & \cellcolor{red!20}Higher & \cellcolor{green!20}Lower & \cellcolor{green!20}Lower & \cellcolor{yellow!20}Medium \\
\bottomrule
\end{tabular}
\caption{Comparison of networking infrastructure designs.
}
\label{tab:networking_comparison}
\vspace{-6mm}
\end{table}

\myparagraph{Rearchitecting for AI} 
AI workloads impose far greater network demands than general-purpose datacenters. 
Emerging practice starts adopting hierarchical designs for AI inference workloads to match communication patterns~\cite{li2025llmhardware}: 
NVLink~\cite{nvlink} for intra-server to support tensor parallelism (TP)~\cite{shoeybi2019megatron}, and lower-cost networks for pipeline parallelism~\cite{shoeybi2019megatron}, which exchanges data less frequently and is employed across servers. 
We evaluate four network designs: 
(1) all-accelerator connectivity over Ethernet, 
(2) all over InfiniBand~\cite{infiniband}, 
(3) all over NVLink~\cite{nvlink}, and 
(4) a \emph{hierarchical} approach (NVLink within a server, InfiniBand within a rack, and Ethernet across racks). 
\Cref{tab:networking_comparison} shows 
different cost–performance trade-offs across networking strategies.
\Cref{fig:tco_networking_infra} shows that the hierarchical design delivers the best balance, reducing TCO by 6\% relative to a flat high-performance network to achieve the workload latency requirements. By aligning interconnect performance with workload communication patterns, hierarchical networking ensures scalability without over-provisioning expensive, low-latency links.

\subsection{Lessons}
AI workloads fundamentally challenge legacy datacenter designs, creating unprecedented demands on power, cooling, and networking.
As accelerator power densities increase and communication patterns grow more complex, traditional hierarchical power distribution, air-based cooling, and uniform network fabrics become increasingly inadequate; both in terms of cost efficiency and workload performance.
These rapidly evolving requirements have spurred a wave of infrastructure innovations.
By evaluating design choices from the ground up using our TCO-driven framework over the full datacenter lifecycle, we show how emerging solutions (\eg{}, flatter power delivery architectures, hybrid cooling systems, and hierarchical network designs) can meet long-term efficiency and cost objectives.
\section{Provisioning AI Hardware}
\label{sec:stage_refresh}

Once the datacenter infrastructure is built, the next step is managing the hardware lifecycle:
deciding when and how to retire outdated components and introduce new ones.
Careful refresh planning ensures the system continues to meet performance, efficiency, and scalability demands.

\subsection{Current Hardware AI Trends}
\label{sec:current_hw_ai_trends}

\begin{figure}[t]
    \centering
    \includegraphics[width=\linewidth]{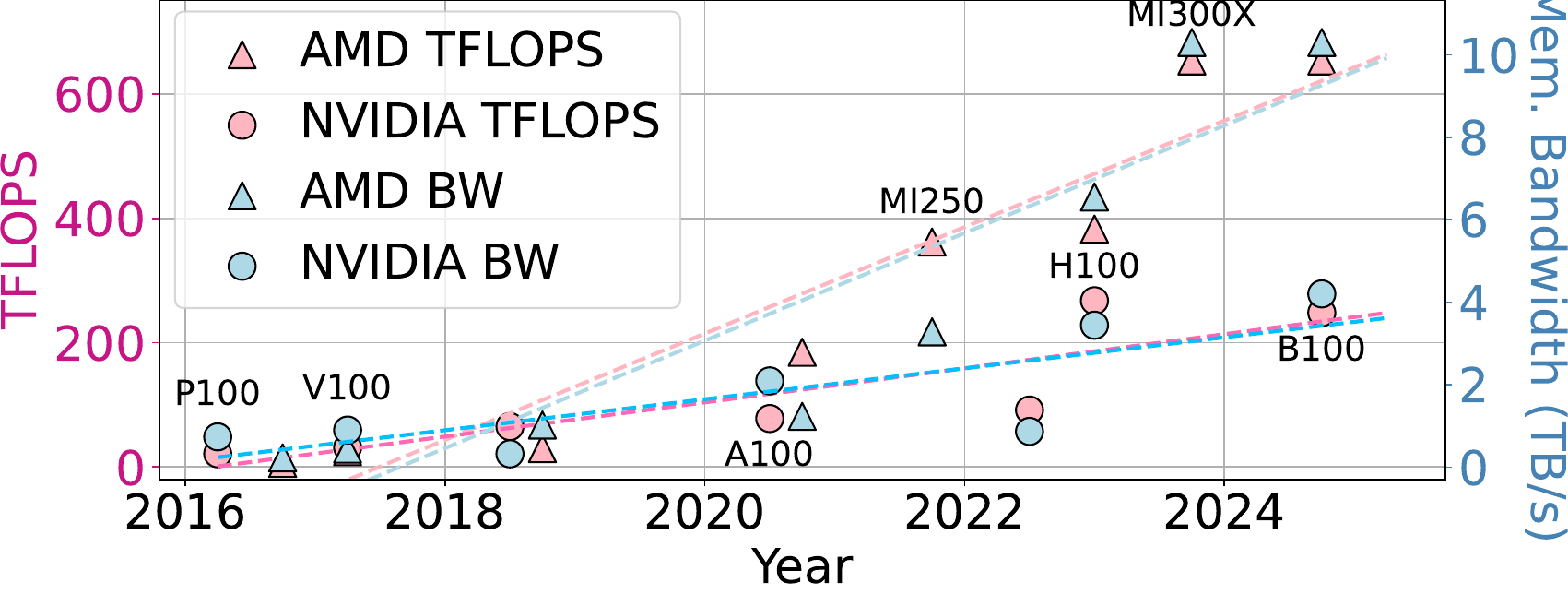}
    \vspace{-6mm}
    \caption{Evolution of AMD and NVIDIA GPUs showing TFLOPS (left axis) and memory bandwidth (right) over time. 
    }
    \label{fig:nvidia_vs_amd}
    \vspace{-3mm}
\end{figure}

\myparagraph{GPU Release Cycle}
GPUs from vendors like NVIDIA and AMD have evolved significantly.
Starting with the NVIDIA P100 (Pascal), which offered modest FP16/FP32 throughput, modern GPUs such as the B200 deliver vastly higher performance through advanced tensor cores, increased memory bandwidth, and optimizations in sparsity and mixed precision.
\Cref{fig:nvidia_vs_amd} shows the progression of compute capability (TFLOPS) and memory bandwidth of NVIDIA datacenter GPUs from 2016 to 2025.
Compute throughput has increased nearly 12\texttimes{}, while memory bandwidth (critical for large-scale and sparse model inference) has grown over 5\texttimes{}.
These trends are even more pronounced for AMD GPUs~\cite{amdMI}. 

GPU vendors are aggressively releasing new architectures every year 
(even multiple generations per year)
to meet the rapidly evolving demands of AI.
This pace contrasts the traditional 2--3 year cadence of CPU releases.
Similar trends hold for other specialized AI hardware such as TPUs~\cite{tpuv4} and LPUs~\cite{groqlpu}, which also follow fast-paced release cycles. 

\myparagraph{Cost}
The cost of GPU servers has also grown substantially.
For example, the NVIDIA P100 was priced at around \$9K per GPU (roughly \$90K per server), whereas the modern H100 costs about \$30K per GPU (over \$350K per server)~\cite{h100Cost}.
AMD’s MI series accelerators follow a similar trend, with newer generations carrying significantly higher price tags.
In contrast, CPU server costs have risen more moderately over the same period.
For example, from around \$7K per server for Intel’s Haswell~\cite{haswellCost} systems in 2014 to approximately 
\$12K for the latest Granite Rapids~\cite{graniteCost} generation in 2025.

\myparagraph{Performance}
We evaluate how the performance of AI inference workloads with different model sizes and architectures scales across GPU generations. We ran experiments using real hardware on vLLM~\cite{vllm} and compare against the roofline model, which shows within 5\% of errors for Llama3~\cite{llama3} models. 
All runs use a 2K sequence length and batch size of 8.
We measure TTFT and TBT in milliseconds, normalized to the H200 baseline per model.
We also combine performance with cost and power (\ie{}, Perf/\$ and Perf/Watt).

\begin{figure}[t]
    \centering
    \includegraphics[width=\linewidth]{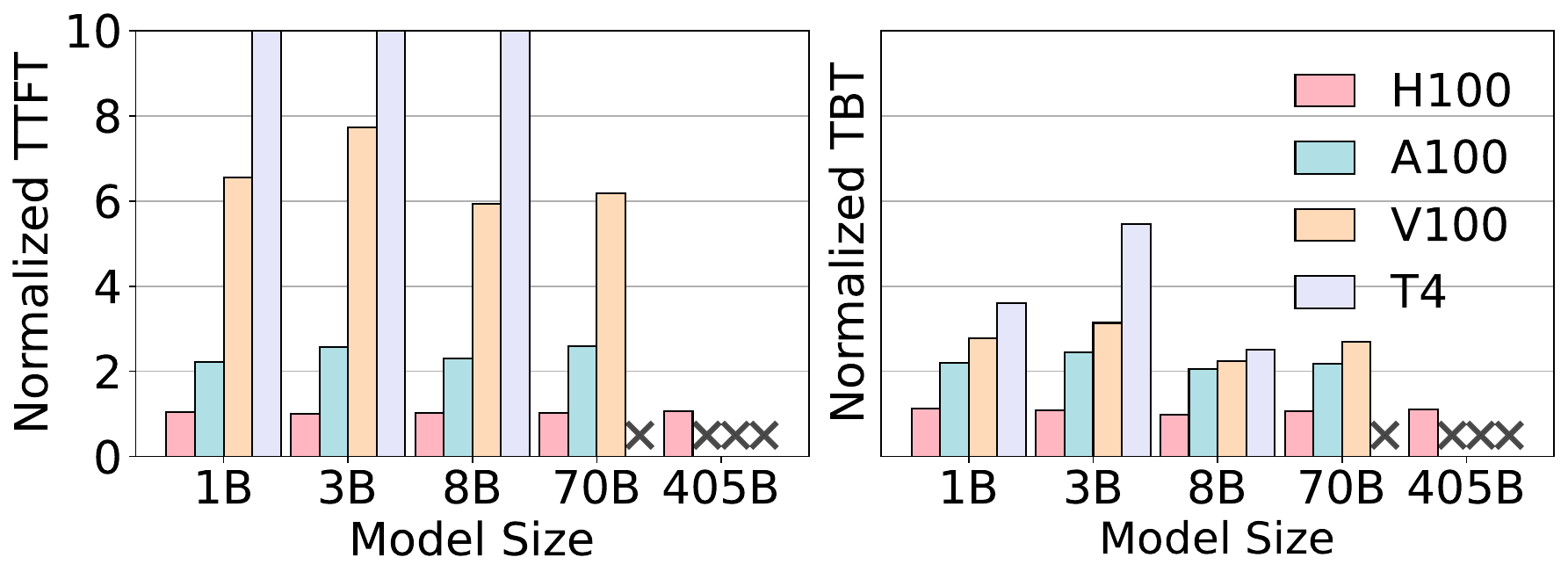}
    \vspace{-6mm}
    \caption{TTFT and TBT latencies for different sizes of Llama-3 LLM across GPU generations normalized to H200. 
    }
    \label{fig:modelsize}
    \vspace{-3mm}
\end{figure}

\myparagraphemph{Model Size}
We evaluate scalability using Llama3~\cite{llama3} models from 1B to 405B parameters on NVIDIA GPUs:
T4, V100, A100, H100, and H200.
We configure TP~\cite{shoeybi2019megatron} with the smallest value that fits each model across most GPUs (\eg{}, TP1 for 1B/3B, TP4 for 8B, TP8 for 70B/405B). Some setups fail on older GPUs due to memory limits.
\Cref{fig:modelsize} shows that older GPUs remain viable for small models.
The decode phase (TBT) is less sensitive to GPU generation than the prefill phase, reflecting its lower compute demands~\cite{splitwise}.

\begin{figure}[t]
    \centering
    \includegraphics[width=\linewidth]{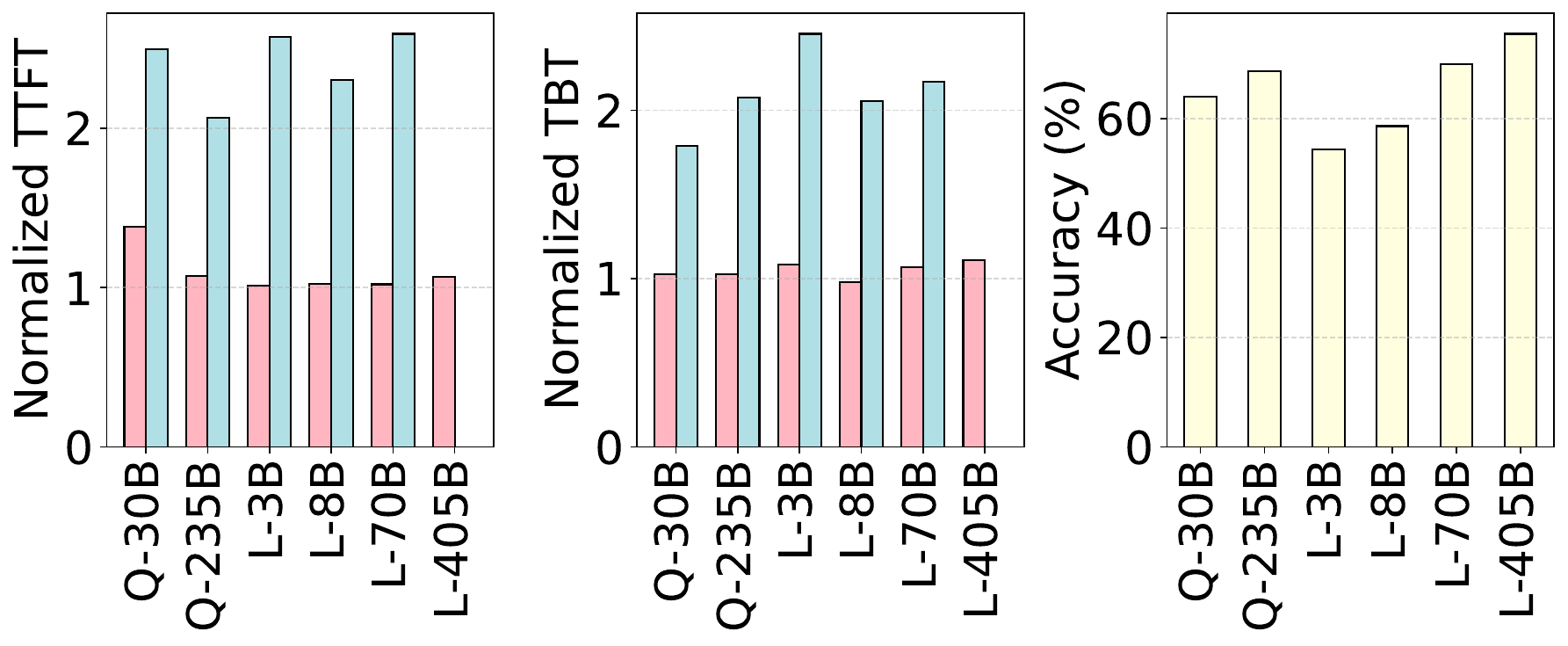}
    \vspace{-5mm}
    \caption{Latencies and accuracies for dense (Llama3) and sparse models (Qwen3) across GPU generations normalized to H200. 
    Pink and blue bars for H100 and A100, respectively.}
    \label{fig:model_sparsity}
    \vspace{0mm}
\end{figure}

\myparagraphemph{Model Architecture}
We evaluate the impact of model sparsity by comparing sparse Qwen3 models (30B A3B and 235B A22B) with dense Llama3 models of similar sizes.
\Cref{fig:model_sparsity} shows that sparse models scale better on older GPUs, maintaining competitive accuracy while outperforming dense counterparts in latency.
For example, Qwen3-235B-A22B matches Llama3-70B in accuracy but degrades less on older hardware (though it requires nearly twice the GPU memory).
This highlights the value of sparsity-aware designs for extending the utility of legacy hardware.

\begin{figure}[t]
    \centering
    \includegraphics[width=\linewidth]{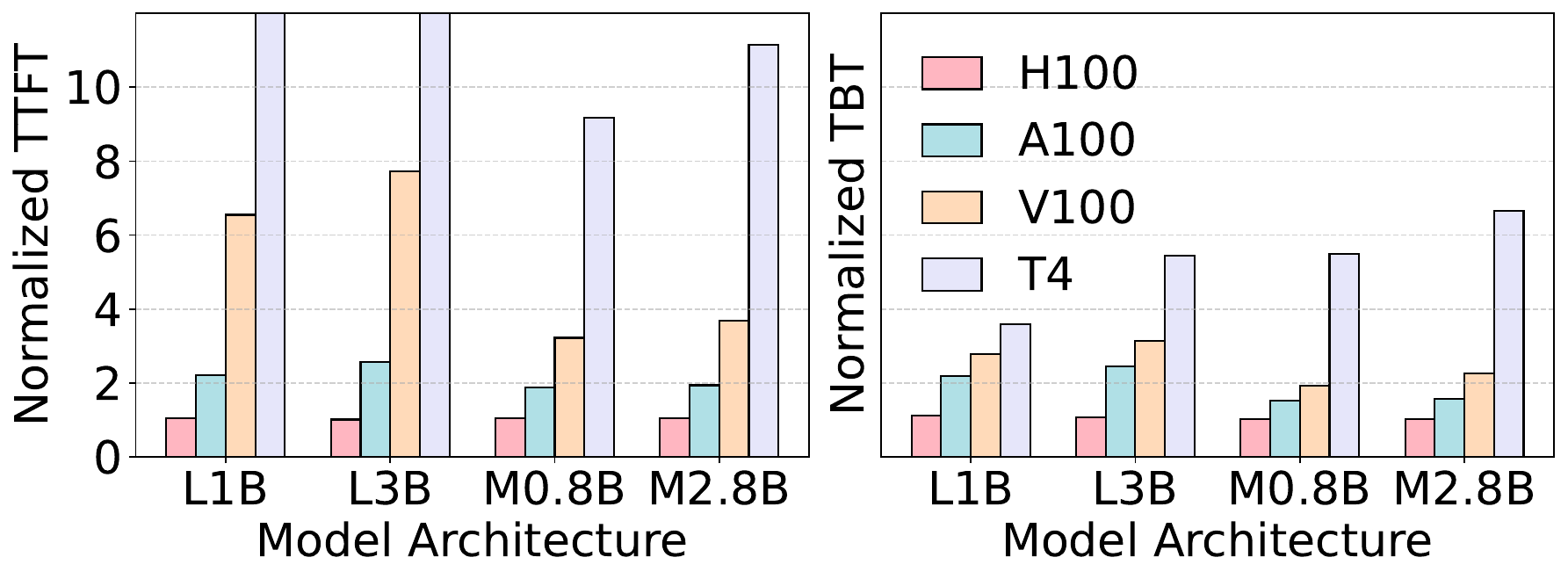}
    \vspace{-5mm}
    \caption{Latencies for transformer (Llama3) and state space (Mamba) models across GPUs normalized to H200.}
    \label{fig:model_arch}
    \vspace{-3mm}
\end{figure}

\Cref{fig:model_arch} compares transformer-based (Llama3-3B) with state-space-based (Mamba-2.8B).
State-space models are more hardware-efficient:
for 2K sequences using TP1, Llama3 runs 7.7\texttimes{} slower on V100 than on H200, while Mamba slows down by only 3.6\texttimes{}.
This shows the architectural compatibility of state-space models with older or less performant GPUs.

\myparagraphemph{Model Efficiency}
We combine model performance and hardware cost and power usage to evaluate model efficiency on different generations of hardware.
While the goodput per Watt of large models degrades substantially on older hardware, the gap is much smaller for smaller models.
For instance, running Llama3-70B on an A100 yields roughly 3$\times$ lower goodput per Watt compared to an H100, whereas running Llama-1B on an A100 is only about 18\% lower. In terms of goodput per Watt per dollar, the A100 actually outperforms the H100 by 8–23\% for the 1B, 3B, and 8B models, but delivers about 2$\times$ lower efficiency for Llama-70B.
In fact, for smaller models, even the V100 is on par with the H100, achieving performance per Watt per dollar within 5\% of it. 

\subsection{Refresh Policies}

\myparagraph{Traditional Approach}
General-purpose datacenters typically follow a steady CPU refresh cycle, with servers remaining in service for about five years before replacement~\cite{wang2024designing}.
This baseline policy strikes a balance between capital and operational efficiency.

Alternative strategies include extending server lifetimes to reduce CapEx 
(at the expense of higher energy use and maintenance) or shortening lifetimes to deploy more efficient hardware sooner, increasing capital costs but improving energy and space efficiency.
Skipping intermediate generations is another option when current hardware meets workload needs and newer gains are marginal.

\begin{figure}[t]
\centering
\subfloat[General-purpose datacenters.]{
  \includegraphics[clip,width=0.5\columnwidth]{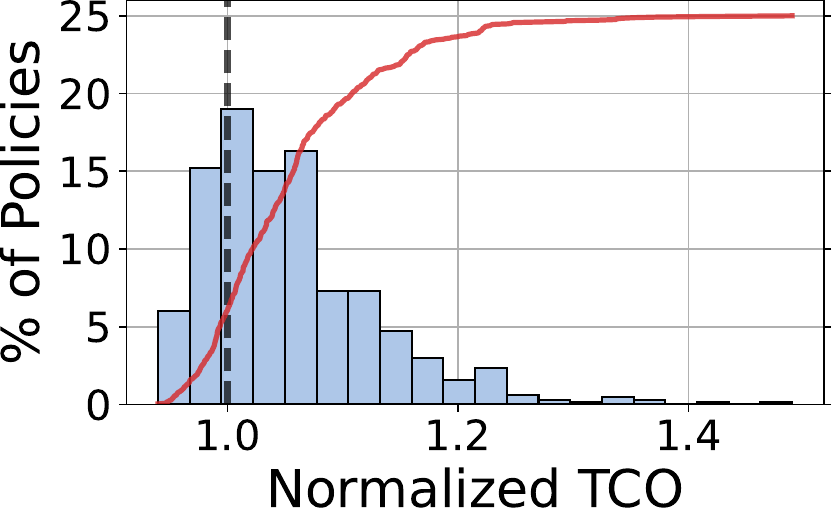}
  \label{fig:refresh_gp}
}
\subfloat[AI datacenters.]
{
  \includegraphics[clip,width=0.5\columnwidth]{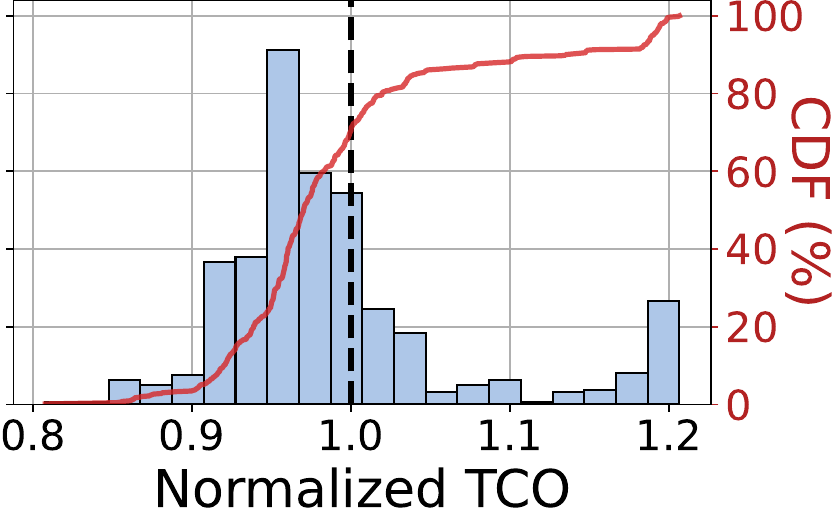}
  \label{fig:refresh_gpu}
}
\vspace{-3mm}
\caption{TCO distribution for various hardware refresh policies in general-purpose and AI datacenters.
}
\label{fig:refresh}
\vspace{0mm}
\end{figure}

We use our framework to evaluate the policies to provision and refresh the servers in an AI datacenter.
\Cref{fig:refresh_gp} shows the TCO distribution of all feasible refresh policies normalized to baseline policy value.
Most of the policies lie on the right side of the distribution, indicating the 5-year baseline remains among the most cost-effective strategies for general-purpose datacenters. 

\begin{figure}[t]
    \includegraphics[width=\columnwidth]{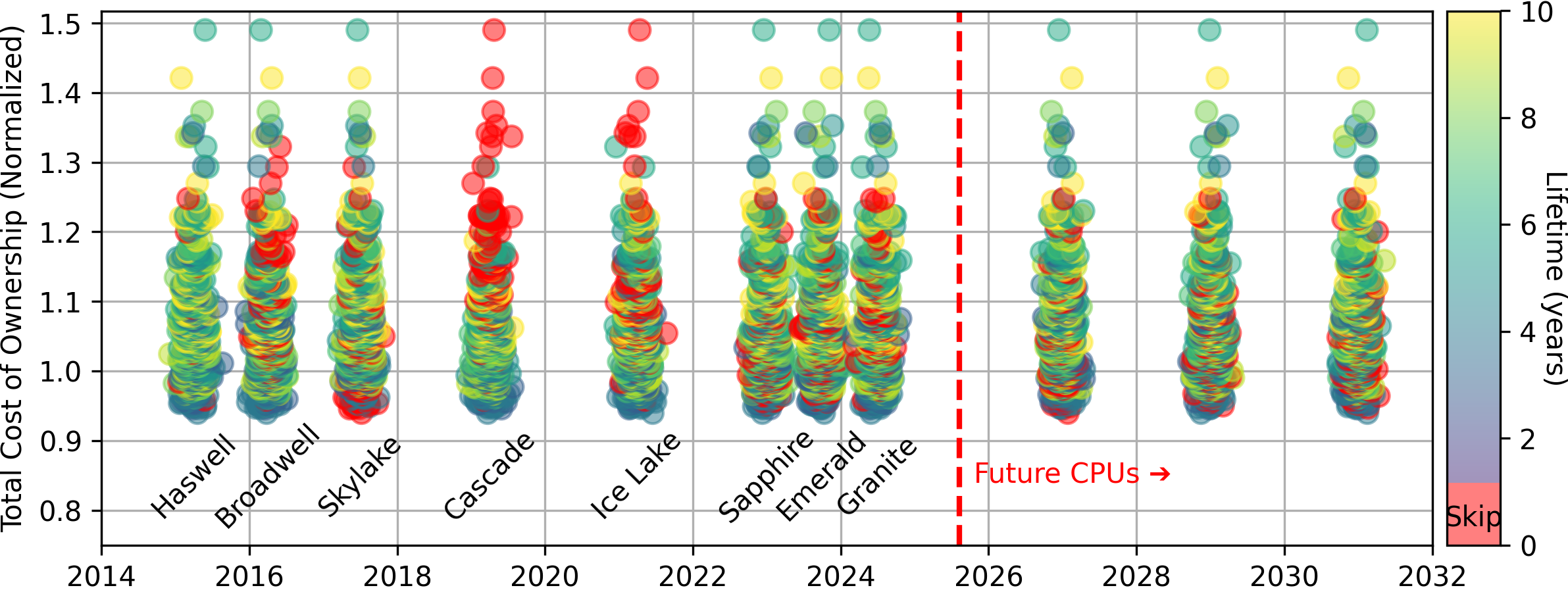}
    \includegraphics[width=\columnwidth]{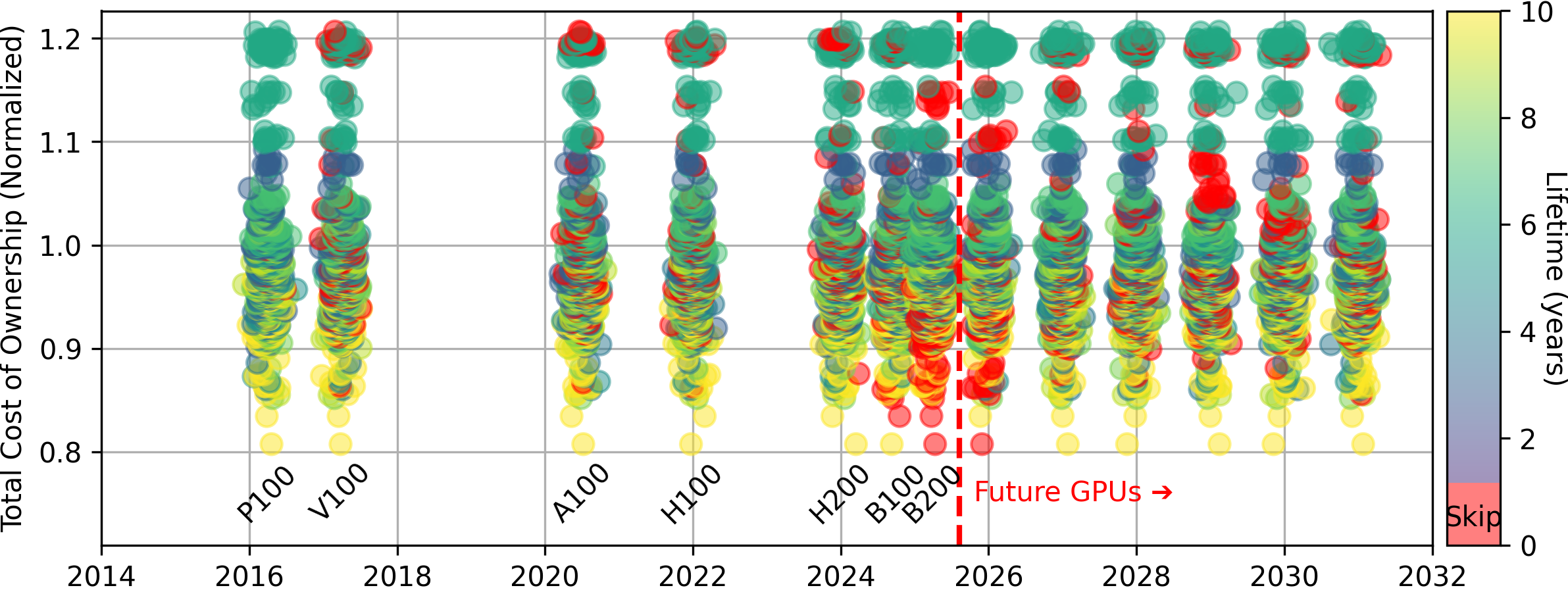}
    \vspace{-6mm}
    \caption{TCO changing refresh policy for each hardware generation in general-purpose and AI datacenters. 
    }
    \label{fig:refresh_year}
    \vspace{-4mm}
\end{figure}

\Cref{fig:refresh_year} shows the same data split by hardware generation.
The top of \Cref{fig:refresh_year} presents a per-CPU-generation view, showing how varying the refresh policies of each individual hardware generation, from 0 years (skipping that generation entirely) to 10 years, impacts normalized TCO.
For general-purpose datacenters, except for Skylake (where skipping was preferable), the refresh policies of 4--6 years lifetime have the lowest TCOs. 

\myparagraph{Rearchitecting for AI}
The dynamics of AI accelerators and workloads differ substantially from those of general-purpose datacenters,
rendering the traditional refreshes sub-optimal. 
\Cref{fig:refresh_gpu} shows that many alternative refresh strategies can reduce TCO by 15--20\% compared to the baseline.

\myparagraphemph{Hardware Trend}
\Cref{fig:refresh_year} shows the TCO changing the refresh cycle (and skipping) for each hardware generation.
Three dominant hardware-driven trends emerge.
First, \emph{newer is much better}: 
when new GPUs provide substantial efficiency gains, early decommissioning of older hardware is justified, and datacenters benefit from upgrading as soon as the next generation is available. 
For example, transitioning from NVIDIA V100 to A100 GPUs is beneficial.

Second, \emph{older is competitive}:
The generation breakdown shows that for most cases extending their hardware lifespan beyond 6 years is cost-effective.
This was the case for all generations before B100 GPUs.

Third, \emph{newer is similar or worse}: when new GPUs offer marginal or negative efficiency gains, it is better to extend the life of existing hardware and skip intermediate generations.
For example, it is more cost effective to skip B100/B200 GPUs.

\myparagraphemph{Workload Evolution}
If models grow significantly year over year (as seen in past GPT-family releases~\cite{openAIModels}) refresh cycles must be accelerated to provision more efficient hardware that can handle the increased compute needs.
Conversely, if model sizes stabilize or decrease, extending hardware lifetime becomes more cost-effective.

\begin{figure}[t]
    \includegraphics[width=\columnwidth]{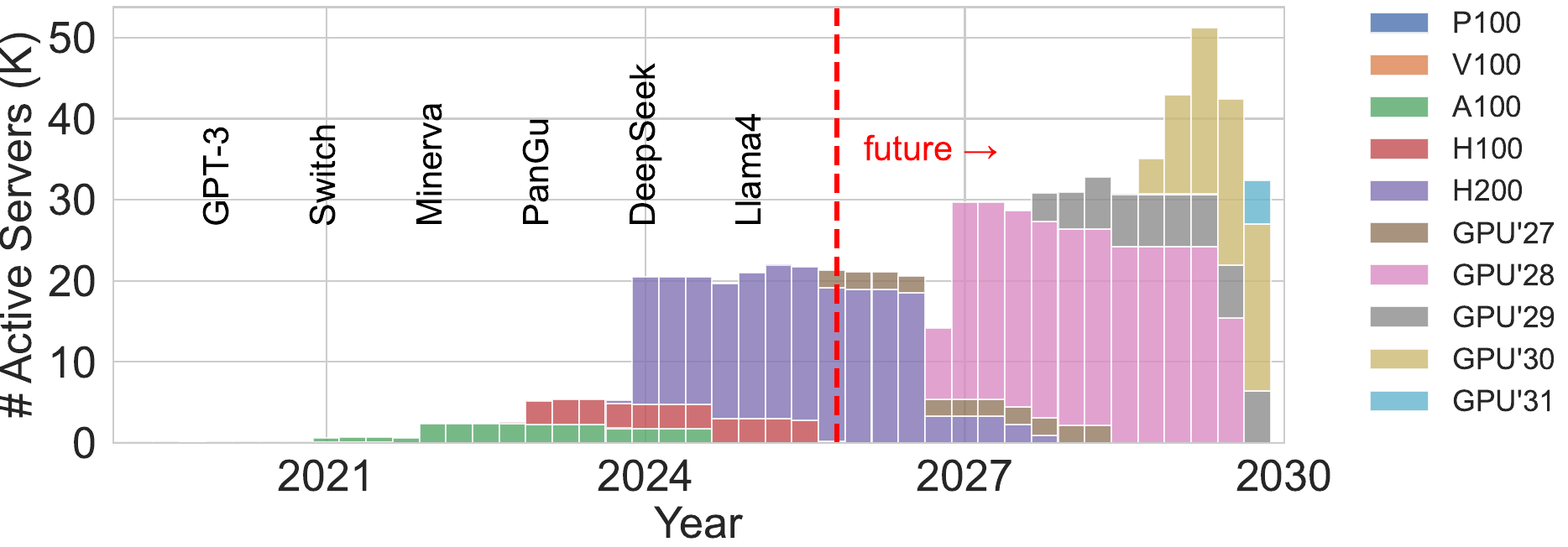}
    \vspace{-7mm}
    \caption{Server count by GPU type over time in an AI fleet following the optimal refresh policy for minimizing TCO.
    }
    \label{fig:server_count_opt}
    \vspace{0mm}
\end{figure}

\myparagraphemph{Example Timeline}
\Cref{fig:server_count_opt} shows the deployment of GPU server generations under the optimal refresh policy to minimize TCO.
The policy skips some generations (\eg{}, B100, B200) entirely, as extending the service life of H100 and H200 GPUs proves more cost-effective.
When a newer generation delivers substantial performance and efficiency gains, the policy triggers earlier decommissioning and demand-driven purchases to match workload growth and model sizes.
Compared to the baseline policy in \Cref{fig:server_count_baseline}, this approach yields a smoother, more balanced mix of old and new hardware, and at times even a modest reduction in total server count due to improved GPU efficiency (\eg{}, in early 2027).

\subsection{Lessons}
Fixed refresh intervals are no longer sufficient for AI workloads.
Unlike traditional datacenters, where hardware efficiency and workload demands change gradually, AI accelerators and models evolve at a much faster pace.
Some generations deliver dramatic performance gains, while others bring only marginal improvements.
These shifts are further amplified by rapid increases in power and thermal densities, hardware costs, and release frequency, changing the trade-offs between raw performance and efficiency.

Therefore, AI datacenters must adopt flexible hardware refresh strategies that respond to evolving hardware efficiency and workload trends.
This may involve aggressively retiring older GPUs when new generations deliver significant efficiency gains, while extending the life of existing hardware (or skipping intermediate generations) when improvements are limited.
By aligning refresh decisions with model scaling and operational cost trends, infrastructure can evolve more intelligently, maximizing performance while controlling costs.

\section{Operating an AI Datacenter}

Once hardware is provisioned, the challenge becomes sustaining high utilization while meeting SLOs across a diverse, evolving fleet. 
This demands software techniques to orchestrate diverse workload types, hardware generations, 
and different performance-cost trade-offs, all shaped by earlier building and refreshing decisions.

\begin{figure}[t]
    \centering
    \includegraphics[width=\linewidth]{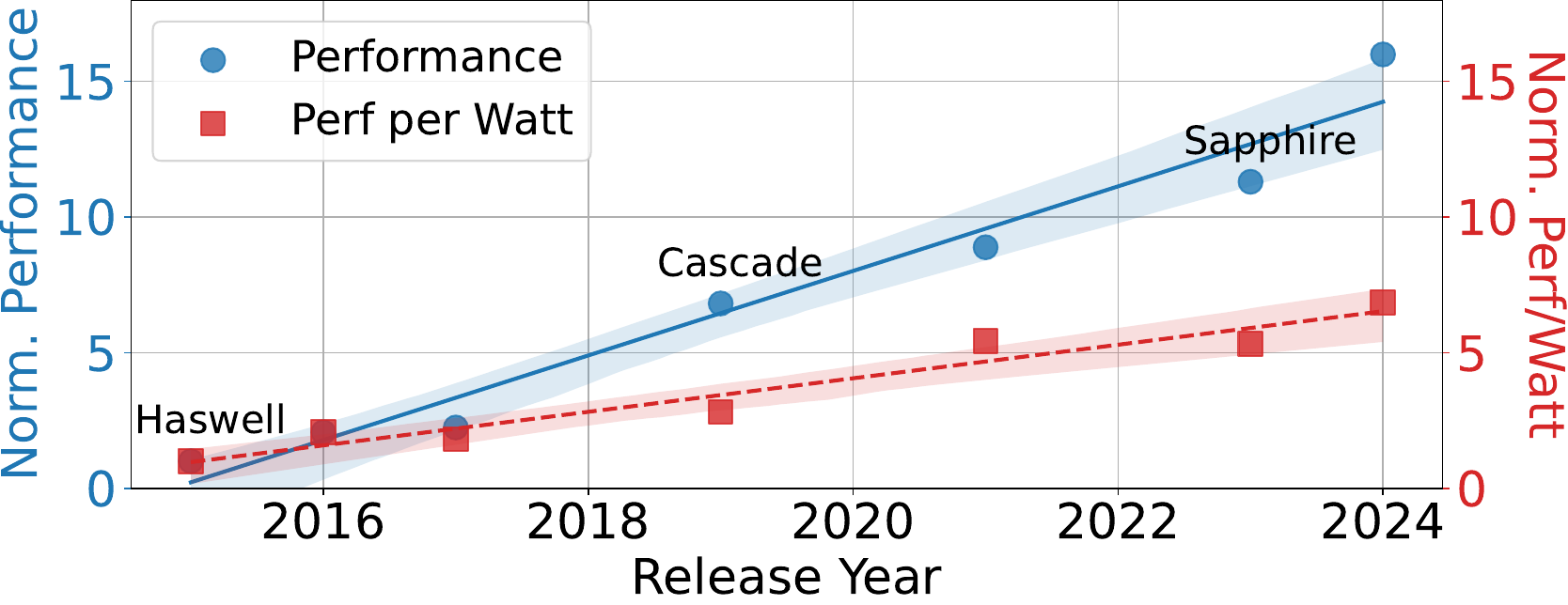}
    \vspace{-6mm}
    \caption{Performance and performance/Watt for DCPerf applications~\cite{dcperf} across generations of Intel servers.}
    \label{fig:dc_perf}
    \vspace{0mm}
\end{figure}

\myparagraph{Traditional Approach}
In general-purpose datacenters, workloads typically run on homogeneous hardware pools with uniform deployment strategies, from bare-metal servers and VMs to microservices and serverless.
Most workloads are generally deployed on the latest hardware, while some workloads may remain on legacy systems until decommissioned~\cite{migrationHP}.
Software stacks are often tuned for predictable, stable performance, 
requiring little adaptation to hardware diversity or rapid workload changes.

To quantify this, we use the DCPerf benchmark suite~\cite{dcperf}, which includes representative production-grade applications (\eg{}, key-value stores, databases, and video processing) to evaluate performance and power efficiency across multiple Intel server generations.
\Cref{fig:dc_perf} shows aggregated performance and performance per Watt, indicating that both throughput and power efficiency scale nearly linearly with newer servers.
This justifies the common practice of migrating workloads to the latest generation.

\begin{table*}[t]
\footnotesize
\begin{tabular}{lll}
\toprule
\textbf{Optimization Technique} & \textbf{Description}                & \textbf{TCO Impact}        \\
\midrule
\textbf{Smooth Model Migration}~\cite{distilation1,distilation2}                            & Gradual migration from old to newer models upon releases & Avoid rapid hardware procurement \\
\textbf{Model Quantization}~\cite{zhao2024atom, lin2024qserve}                                & Lower precision to reduce compute/memory                 & Lower hardware demand and cost per inference \\
\textbf{KV-Cache Management}~\cite{nvidiaDynamo, gao2024cost}                         & Optimize storage and reuse of KV cache                   & Increase older hardware reuse \\
\textbf{Disaggregation}~\cite{splitwise, zhong2024distserve, nvidiaDynamo, step3, megascale}                     & Split distinct phases onto different hardware       & Extend useful life of heterogeneous generation \\
\textbf{Alternative Architectures} & Mixture-of-Experts~\cite{shazeer2017outrageouslylargeneuralnetworks}, State-Space-Models~\cite{gu2022efficientlymodelinglongsequences} & Increase older hardware reuse\\
\textbf{Model Routing}~\cite{jain2024intelligent, ding2024hybrid}                               & Direct workloads to the most efficient model variant    & Increase older hardware reuse \\
\textbf{Heterogeneity-Aware Scheduling} ~\cite{mei2025helix, jiang2023hexgen}            & Map workloads to optimal/available hardware generation         & Defer refresh costs\\
\textbf{Infrastructure-Aware Scheduling} ~\cite{stojkovic2025tapas, stojkovic2024dynamollm}            & Exploit headroom within infra capacity envelopes         & Improve infrastructure efficiency\\
\bottomrule
\end{tabular}
\vspace{2mm}
\caption{Operation stage software optimizations that introduce new cross-stage optimization opportunities.
}
\label{tab:software-optimization}
\vspace{0mm}
\end{table*}

\myparagraph{Rearchitecting for AI}
As discussed in \Cref{sec:current_hw_ai_trends}, AI workloads and hardware trends diverge significantly from those of general-purpose datacenters. 
Unlike CPUs, we do not observe uniform or linear performance improvements of AI inference workloads 
across GPU generations, 
and the benefits of new hardware vary considerably across different models and use-cases. 
Hence, applying the same direct-migration strategy used in traditional datacenters can be suboptimal for AI workloads.
Instead, achieving operational efficiency for AI requires software optimizations that bridge the gap between 
evolving models, growing user demands, and a heterogeneous fleet, while controlling cost and sustaining SLOs. 

\begin{figure}[t]
  \centering
  \subfloat[TCO \vs{} software optimization.]{%
  \includegraphics[clip,width=0.55\columnwidth]{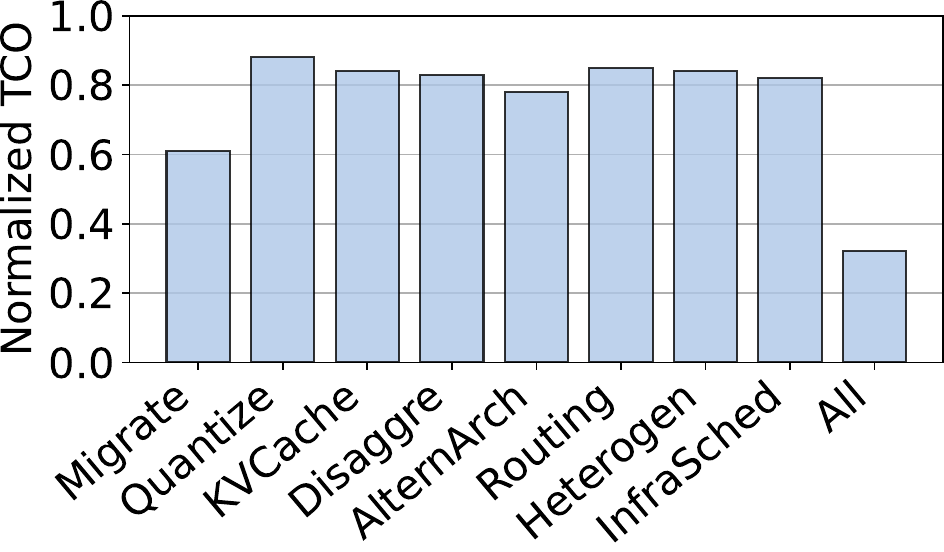}
  \label{fig:operation-tco}
  }
  \subfloat[TCO \vs{} stages. 
  ]{%
  \includegraphics[clip,width=0.45\columnwidth]{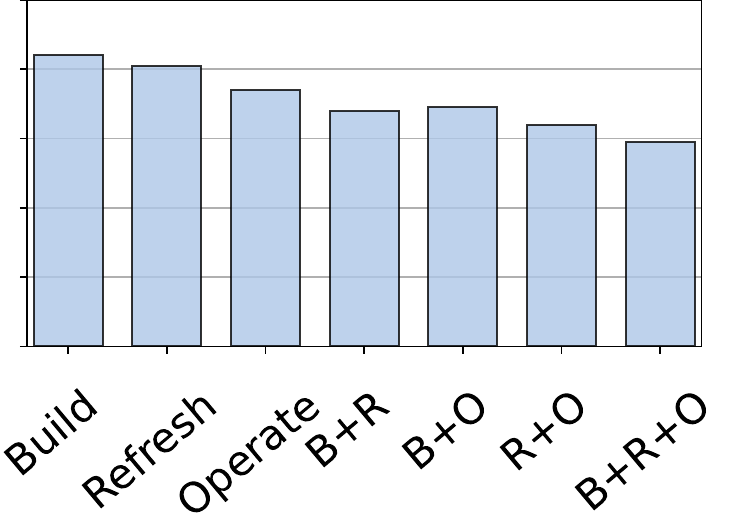}
    \label{fig:stages-tco}}
        \vspace{-3mm}
 \caption{TCO for optimization during \emph{operation} and for stages, normalized to the baseline without optimizations.}
  \vspace{-3mm}
\end{figure}

\myparagraphemph{Optimizations}
\Cref{tab:software-optimization} summarizes some key operation optimizations that influence build and refresh strategies.
\emph{Model migration}, \emph{quantization}, and \emph{KV-cache management} reduce compute and memory pressure.
\emph{Disaggregation} leverages fast interconnects and aligns workload phases with suitable hardware generations. 
\emph{Heterogeneity-aware} routing, placement, and scheduling adapt workloads in real time to match hardware with workload characteristics. 
\emph{Infrastructure-aware scheduling} optimizes for datacenter constraints (\eg{}, power and cooling) via software controls (\eg{}, dynamic sharding), linking operational policies to build-time decisions and enabling lifecycle-wide optimization.

\myparagraphemph{TCO savings}
These optimizations extend the life of older hardware, defer costly refreshes, and improve infrastructure value.
\Cref{fig:operation-tco} shows TCO reductions of 12--39\% per strategy compared to the baseline.
\emph{Smoothen model migration} delivers the largest gains by reducing post-release compute load.
\emph{Disaggregation} and \emph{infrastructure-aware scheduling} achieve strong savings without modifying load, by leveraging heterogeneous fleets and aligning software with datacenter constraints.
Combining \emph{all} optimizations yields over 60\% TCO reduction.
While not fully additive due to overlaps, the cumulative impact is substantial.

\subsection{Lessons}
Fixed, uniform operational strategies are insufficient for rapidly evolving models and heterogeneous hardware generations.
Software-driven techniques (\eg{}, model migration smoothing, disaggregation, and heterogeneity-aware scheduling) shift operations from a reactive process to a proactive, cost-optimized stage of the datacenter lifecycle.
These optimizations extend the useful life of older hardware, leverage heterogeneous fleets, and dynamically align workloads with available infrastructure to capture efficiency gains.
By closing the loop between operations, hardware refresh, and build stages, these techniques turn rigid hardware timelines into adaptive, lifecycle-aware policies.

\section{Cross-Stage Optimizations} 
Optimizing individual stages for AI reduces TCO, but the largest gains come from cross-stage rearchitecting the end-to-end AI datacenter lifecycle.
We outline the current cross-stage strategies and explore emerging opportunities. 

\subsection{Existing Cross-Stage Optimizations}

\myparagraph{IT provisioning \textrightarrow{} Build}
AI hardware trends are reshaping the datacenter build.
Power hierarchies are flattening to support high-density accelerators, cooling is shifting from air to liquid~\cite{blackwellLiquid}, and networking is moving beyond Ethernet to InfiniBand and NVLink fabrics~\cite{nvlink}.
These choices raise initial build costs but ease future refreshes, extending the usable lifetime of each deployment.

\myparagraph{Operate \textrightarrow{} IT provisioning}
Heterogeneity-aware scheduling helps repurposing older GPUs for workloads better suited to their capabilities:
compute-intensive phases (\eg{}, prefill or large models) run on newer GPUs, while memory- or bandwidth-bound phases (\eg{}, decode or smaller models) are offloaded to older generations~\cite{splitwise, mei2025helix}. 
This strategy smooths refresh costs and maintains high utilization, turning hardware upgrades into opportunities for redistribution and continued value rather than premature hardware retirement.

\myparagraph{Build \textrightarrow{} Operate}
Infrastructure decisions made at build time shape operational flexibility. 
Scheduling frameworks translate these choices into software controls that smooth demand, enable safe hardware derating, and sustain efficiency.
Conversely, coordinated derating of servers and power devices within the power hierarchy allows oversubscription and denser deployments; effectively ``upgrading'' infrastructure at runtime without new physical buildouts~\cite{stojkovic2025tapas,polca}.

\myparagraph{Compound TCO Benefits}
\Cref{fig:stages-tco} shows that cross-stage strategies amplify impact across the lifecycle.
Optimizing individual stages yields 20--30\% savings, while combining build and refresh policies pushes savings beyond 35\%.
A holistic approach can reduce TCO by over 40\%. 
Assuming hardware and model sizes grow linearly (center cells in \Cref{tab:cross-stage-streategies}), the best infrastructure to \emph{build} includes flat power delivery, hybrid liquid-air cooling, and a hierarchical networking with Ethernet, Infiniband, and NVLink.
For hardware \emph{refresh}, it is best to extend server lifetimes to five years and adopt new hardware as it becomes available.
During \emph{operation}, the best strategy is to combine \emph{all} available optimizations.

\subsection{Opportunities for Cross-Stage Optimizations}
Looking ahead, several opportunities emerge when infrastructure, hardware, and software are explicitly co-designed with lifecycle interplay in mind.

\myparagraph{Infrastructure}
Today’s software supports heterogeneous fleets, but build and refresh strategies can better leverage heterogeneity.
Rack-level provisioning with mixed-generation accelerators or general-purpose compute reduces interconnect bottlenecks and power fragmentation.
Combined with heterogeneous derating, these setups adapt efficiently to workload demands.
While they require upfront investment in fine-grained telemetry and control, they offer substantial long-term efficiency gains.

\myparagraph{Hardware}
Emerging AI accelerators have traditionally shaped datacenter infrastructure design.
Looking ahead, future accelerators should be designed not only for performance but also for long-term compatibility with the existing infrastructure.
Lower power density and moderated TDP simplify power delivery and cooling, reducing fragmentation.
For example, accelerators could combine high-power SMs to handle the compute-bound prefill phase with Processing-in-Memory (PIM) units tailored for the memory-bound decode phase, enabling more efficient execution within the same server.

\myparagraph{Operation}
Techniques such as KV-cache management or new 
model architectures (\eg{}, MoEs)
shift the balance between compute and memory needs, reshaping both refresh priorities and placement strategies.
Cross-stage planning anticipates these shifts by provisioning memory or storage servers that support multiple GPU generations.

\subsection{Future Trends}

As AI models and hardware keep evolving, lifecycle management must adapt not only to these trends individually but also to their compounded effects.
We analyze model scaling and hardware trajectories to offer guidance for future AI datacenter lifecycles.
\Cref{tab:cross-stage-streategies} shows the optimal strategies across \emph{build}, \emph{IT provisioning}, and \emph{operate} stages, tailored to projected growth patterns in both hardware and models.
We explore \emph{slow} (sub-linear), \emph{medium} (linear), and \emph{fast} (exponential) growth of model size/complexity and hardware.

\begin{table}[t]
\vspace{0mm}
\resizebox{\columnwidth}{!}{%
\begin{tabular}{cc c@{}p{8pt}c@{}p{8pt}c}
\multicolumn{2}{c}{} & \multicolumn{5}{c}{\textbf{Model Growth}} \\
\multicolumn{2}{c}{} & \multicolumn{1}{c}{\textbf{Slow}~\sarrow} && \multicolumn{1}{c}{\textbf{Medium}~\marrow} && \multicolumn{1}{c}{\textbf{Fast}~\farrow}
\\
\multirow{11}{*}{\rotatebox{90}{\textbf{Hardware Growth}}}
& \sarrow & \cellcolor[HTML]{E4ED8E}Per-PDU\quad Air\quad IB && \cellcolor[HTML]{E4ED8E}Per-PDU\quad Air\quad NVLink && \cellcolor[HTML]{E4ED8E}Per-PDU\quad Air\quad NVLink \\
& \marrow & \cellcolor[HTML]{BBC657}Flat\quad Hybrid\quad IB && \cellcolor[HTML]{9DB53E}Flat\quad Hybrid\quad Hierarchy  && \cellcolor[HTML]{BBC657}Flat\quad Liquid\quad Hierarchy\\
& \farrow & \cellcolor[HTML]{CDD962}Flat\quad Liquid\quad IB && \cellcolor[HTML]{BBC657}Flat\quad Hybrid\quad IB     && \cellcolor[HTML]{9DB53E}Flat\quad Hybrid\quad Hierarchy\\
\noalign{\vskip 8pt}
& \sarrow & \cellcolor[HTML]{FEE4C6}8 years\quad Skip    && \cellcolor[HTML]{FEE4C6}8 years\quad Skip        && \cellcolor[HTML]{FEE4C6}8 years\quad Skip \\
& \marrow & \cellcolor[HTML]{F5B866}4 years\quad Buy new && \cellcolor[HTML]{F5D0A0}5 years\quad Buy new     && \cellcolor[HTML]{F5C98F}4 years\quad Skip \\
& \farrow & \cellcolor[HTML]{F5B866}4 years\quad Buy new && \cellcolor[HTML]{F5B866}4 years\quad Buy new     && \cellcolor[HTML]{FF9F1C}3 years\quad Buy new \\
\noalign{\vskip 8pt}
& \sarrow & \cellcolor[HTML]{BD504A}All Optimizations && \cellcolor[HTML]{FFBDBB}Migration + Quantization   && \cellcolor[HTML]{FFBDBB}Migration + Quantization\\
& \marrow & \cellcolor[HTML]{E6836C}Disaggregation    && \cellcolor[HTML]{BD504A}All Optimizations          && \cellcolor[HTML]{FFBDBB}Migration + Quantization\\
& \farrow & \cellcolor[HTML]{CD705A}Disagg. + Hetero. && \cellcolor[HTML]{E6836C}Disaggregation             && \cellcolor[HTML]{BD504A}All Optimizations\\
\end{tabular}%
}
\vspace{2mm}
\caption{Optimal cross-stage strategies based on model and hardware trends under exponential user demand growth.
The rows represent \emph{build}, \emph{refresh}, and \emph{operate} stages.
Color gradient shows degree of lifecycle adaptation required.
}
\label{tab:cross-stage-streategies}
\vspace{-4mm}
\end{table}

\myparagraph{AI Hardware}
When hardware performance scales rapidly, frequent \emph{refresh} cycles are advantageous, justifying earlier adoption and earlier decommissioning.
However, if performance gains slow or power efficiency declines, it becomes more beneficial to extend refresh intervals and design infrastructure for long-term use.
Lower-power, lower-density hardware can also extend infrastructure lifetimes by reducing the need for costly upgrades to power and cooling systems.

\myparagraph{AI Model Size}
So far, we have assumed that AI models will grow linearly.
Larger models amplify compute, memory, and networking requirements, accelerating refresh cadence and demanding infrastructure explicitly designed for modular expansion.
If model size growth slows down, operators can rely on more cost-effective infrastructures (\eg{}, InfiniBand networking). 
Conversely, if models continue to scale rapidly, more expensive infrastructures become necessary (\eg{}, NVLink), 
alongside more aggressive software optimizations (\eg{}, model migration and quantization).

\subsection{Lessons}
Cross-stage optimization reframes lifecycle management as a continuum rather than a sequence.
Build-time choices, refresh strategies, and operation policies compound to drive long-term efficiency.
Looking ahead, future datacenters should be increasingly co-designed across infrastructure, hardware, and software; with lifecycle-aware thinking at the core of scalable, cost-effective AI.

\section{Conclusion}

The rapid growth of AI workloads has outpaced traditional datacenter management. 
AI datacenters need flexible designs, advanced cooling, and software for heterogeneous hardware. 
While stage-specific improvements (build, IT provisioning, operate) help, the biggest gains, up to 40\% TCO reduction,
come from a holistic, cross-stage approach that anticipates workload dynamics and hardware trends, 
enabling cloud providers to scale efficiently and control costs.

\bibliographystyle{ACM-Reference-Format}
\bibliography{ref}

\end{document}